\documentclass[10pt,twocolumn,letterpaper]{article}

\usepackage{iccv}
\usepackage{times}
\usepackage{epsfig}
\usepackage{graphicx}
\usepackage{amsmath}
\usepackage{amssymb}
\usepackage{booktabs}
\usepackage{caption} 

\usepackage{xcolor}
\usepackage{listings}
\usepackage{makecell}
\usepackage{tikz}

\usepackage{textcomp}

\usepackage[accsupp]{axessibility}  
 
\usepackage{mdframed}

\usepackage[pagebackref=true,breaklinks=true,letterpaper=true,colorlinks,bookmarks=false]{hyperref}

\definecolor{gold}{HTML}{FBF2D2}
\definecolor{silver}{HTML}{DDDDDD}
\definecolor{bronze}{HTML}{EED2B8}

\definecolor{goldD}{HTML}{D9AE13}
\definecolor{silverD}{HTML}{909090}
\definecolor{bronzeD}{HTML}{9A5F26}

\definecolor{catGreen}{HTML}{238763}
\definecolor{catBlue}{HTML}{1F70AE}

\definecolor{datasheet}{HTML}{1F70AE}

\newcommand{\medal}[3]{\tikz[baseline=(char.base)]{\node[rounded corners=2pt,fill=#1,draw=#2,inner sep=1.5pt] {#3};}}

\newcommand{\bm}[2]{
    \ifcase#1\or
      {\medal{gold}{goldD}{\textbf{#2}}}
    \or 
      {\medal{silver}{silverD}{#2}}
    \or 
      {\medal{bronze}{bronzeD}{#2}}
    \else 
      #2
    \fi\ignorespaces
}

\newcommand{\textapp}{\raise.17ex\hbox{$\scriptstyle\sim$}}

\newcommand*\circlefill[1]{\tikz[baseline=(char.base)]{
            \node[shape=circle,fill,inner sep=0.5pt] (char) {\color{white} \scriptsize #1};}}
\definecolor{lake}{HTML}{2B4570}
\definecolor{river}{HTML}{00A676}
\definecolor{sea}{HTML}{7A9CC6}
\newcommand{\dtLake}{{\color{lake}\circlefill{L}}}
\newcommand{\dtRiver}{{\color{river} \circlefill{R}}}
\newcommand{\dtSea}{{\color{sea} \circlefill{S}}}

\newcommand*\rectfill[1]{\tikz[baseline=(char.base)]{
            \node[shape=rectangle,fill,inner sep=1.5pt] (char) {\color{white} \scriptsize #1};}}
\definecolor{bbox}{HTML}{9098A8}
\definecolor{semantic}{HTML}{F2A541}
\definecolor{panoptic}{HTML}{BA274A}
\newcommand{\stBox}{{\color{bbox}\rectfill{B}}}
\newcommand{\stSemantic}{{\color{semantic}\rectfill{S}}}
\newcommand{\stPanoptic}{{\color{panoptic} \rectfill{P}}}

\newcommand{\dsUnavail}[1]{\textcolor{gray}{#1}}

\iccvfinalcopy 

\def\iccvPaperID{12536} 

\ificcvfinal\fi

\newcommand\copyrighttext{%
  \footnotesize \textcopyright 2023 IEEE. Personal use of this material is permitted. Permission from IEEE must be obtained for all other uses, in any current or future media, including reprinting/republishing this material for advertising or promotional purposes, creating new collective works, for resale or redistribution to servers or lists, or reuse of any copyrighted component of this work in other works.}
\newcommand\copyrightnotice{%
\begin{tikzpicture}[remember picture,overlay]
\node[anchor=south,yshift=10pt] at (current page.south) {\fbox{\parbox{\dimexpr\textwidth-\fboxsep-\fboxrule\relax}{\copyrighttext}}};
\end{tikzpicture}%
}

\begin{document}

\title{LaRS: A Diverse Panoptic Maritime Obstacle Detection Dataset and Benchmark}


\author{Lojze Žust, Janez Perš, Matej Kristan\\
University of Ljubljana\\
{\tt\small \{lojze.zust,matej.kristan\}@fri.uni-lj.si, janez.pers@fe.uni-lj.si}
}





\twocolumn[{%
\renewcommand\twocolumn[1][]{#1}%
\copyrightnotice
\maketitle
\begin{center}
  \centering
  \captionsetup{type=figure}
  \includegraphics[width=\linewidth]{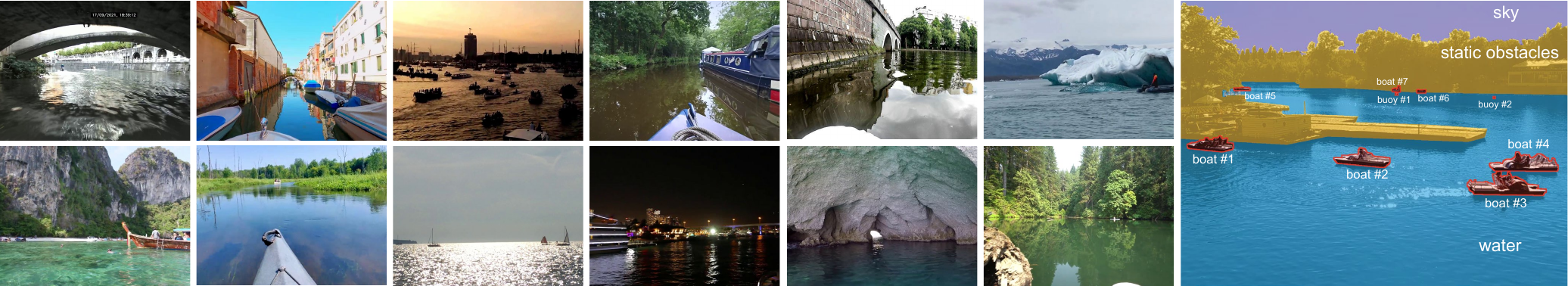}
  \captionof{figure}{LaRS features diverse and challenging USV-centric scenes with per-pixel panoptic annotations (right).}
    \label{fig:motivation}
\end{center}
}]

\ificcvfinal\thispagestyle{empty}\fi


\begin{abstract}
The progress in maritime obstacle detection is hindered by the lack of a diverse dataset that adequately captures the complexity of general maritime environments. We present the first maritime panoptic obstacle detection benchmark LaRS, featuring scenes from \underline{La}kes, \underline{R}ivers and \underline{S}eas. Our major contribution is the new dataset, which boasts the largest diversity in recording locations, scene types, obstacle classes, and acquisition conditions among the related datasets. LaRS is composed of over 4000 per-pixel labeled key frames with nine preceding frames to allow utilization of the temporal texture, amounting to over 40k frames. Each key frame is annotated with 8 thing, 3 stuff classes and 19 global scene attributes. We report the results of 27 semantic and panoptic segmentation methods, along with several performance insights and future research directions. To enable objective evaluation, we have implemented an online evaluation server. The LaRS dataset, evaluation toolkit and benchmark are publicly available at: \url{https://lojzezust.github.io/lars-dataset}
\end{abstract}

\section{Introduction}



The maritime industry is undergoing a fundamental transformation.
With over 90\% of goods being moved over water, substantial efforts are being invested in development of autonomous unmanned surface vessels (USV)~\cite{DeFilippo2021RoboWhaler,Cheng2021Are}. These autonomous boats serve a wide range of purposes, ranging from automated inspection, environmental monitoring, waste cleanup, cargo shipping, to civilian transportation. The autonomy of USVs critically depends on obstacle detection capability for timely collision avoidance. 
Similarly to the automotive domain~\cite{Cordts2016Cityscapes,Geiger2012Are}, cameras 
have been extensively explored for this task~\cite{Kristan2016Fast,Cane2019Evaluating,Prasad2019Object,Bovcon2018Stereo,Nirgudkar2021Visible,Bovcon2021WaSR}.

There are several challenges associated with maritime obstacle detection. The appearance of the navigable surface (water) is dynamic and reflects the environment, often containing strong mirroring and sun glitter (Figure~\ref{fig:motivation}). 
Although modern detectors~\cite{Tan2019EfficientDet,Tian2019FCOS,Carion2020EndtoEnd} can accurately detect common dynamic obstacles such as ships and boats, the appearance of obstacles such as buoys, people and animals can vary significantly, bringing the task closer to anomaly detection~\cite{Lis2019Detecting,Chan2021SegmentMeIfYouCan}. Furthermore, background static obstacles, such as shorelines and piers, cannot be addressed by these methods. 

The currently dominant approach~\cite{Kristan2016Fast,Bovcon2021WaSR} instead employs semantic segmentation to decompose the scene into three semantic classes (water, obstacles and sky), which jointly address static and dynamic obstacles. Nevertheless, the recent detection benchmark~\cite{Bovcon2020MODS} indicates that segmentation methods could benefit from the detection approach. A natural approach that combines these two principles is panoptic segmentation~\cite{Kirillov2019Panoptic}, which has proven highly effective in the related field of autonomous ground vehicles~\cite{Cordts2016Cityscapes,Geiger2012Are,Cheng2021Mask2Former,Zendel2022Unifying}. Unfortunately, panoptic segmentation has not been fully explored for maritime perception, primarily due to the lack of a diverse, publicly available, curated panoptic dataset.
 
Several maritime
evaluation~\cite{Prasad2017Video,Bovcon2018Stereo,Bovcon2020MODS} and training~\cite{Bovcon2019Mastr,Cheng2021Are} datasets have been proposed, as shown in Table~\ref{tab:datasets}. However, a common drawback of the major evaluation datasets is that the dynamic obstacles are annotated only with bounding boxes, limiting the evaluation capability.
Additionally, the current segmentation training datasets~\cite{Bovcon2019Mastr,Cheng2021Are} are modest in size and diversity, and the only reported RGB-based maritime panoptic dataset~\cite{Qiao2022Automated} is private and cannot be utilized by the community.
Moreover, the scene diversity in individual datasets is fairly low, since they are all captured in limited geographic locations, which hampers the development of robust maritime obstacle detection methods capable of handling general maritime environments.

We address the aforementioned drawbacks by proposing the first maritime panoptic obstacle detection benchmark. Our major contribution is the \underline{La}kes \underline{R}ivers and \underline{S}eas (LaRS) dataset (see Figure~\ref{fig:motivation}). 
LaRS surpasses existing datasets in terms of diversity, obstacle types and acquisition conditions. The dataset is composed of over 4000 key frames with panoptic labels for 3 stuff and 8 thing categories, and 19 global scene attributes. Each key frame is equipped with the preceding nine frames to facilitate the development of methods that exploit temporal texture.
To ensure equal attribute distribution, the training, validation, and test splits were carefully constructed, and we have implemented an online evaluation server to mitigate test-set overfitting.
 
In addition to the LaRS dataset, our second contribution is the analysis of 19 recent semantic segmentation networks and 8 panoptic segmentation networks. We highlight several limitations of these methods and identify opportunities for their improvement.
%
The dataset, benchmark, and evaluation toolkit will be publicly released, to enable the research community to utilize and build upon our work.

\section{Related Work}

\paragraph{Maritime obstacle detection.}

The early works in camera-based obstacle detection include statistical semantic segmentation methods~\cite{Kristan2016Fast}, handcrafted saliency estimation~\cite{Cane2016SaliencyBased}, background subtraction~\cite{Prasad2019Object} and stereo reconstruction~\cite{Wang2013Stereovision,Muhovic2020Obstacle}. These methods, however, typically fail on mirroring, glitter and other visual ambiguities. The general-purpose CNN-based object detectors~\cite{Ren2017Faster,Ma2020Convolutional,Bovcon2020MODS} have shown a much better resilience, but do not cope well with long-tail distribution object types and cannot address background static obstacles.

The current dominant line of research stems from the early statistical method~\cite{Kristan2015Graphical}, which proposed segmenting the scene into navigable and non-navigable regions (i.e., water and obstacles), thus jointly addressing dynamic and static obstacles. Several works~\cite{Cane2019Evaluating,Bovcon2019Mastr} have shown that semantic segmentation networks from the AGV domain underperform in the maritime setup and a number of maritime-specific segmentation networks have been proposed since, most notably~\cite{Steccanella2020,Bovcon2021WaSR,Chen2021WODIS,Yao2021Shoreline}. A recent work~\cite{Zust2022Temporal} proposed exploiting the temporal texture to address reflections, while several works considered alternative visual modalities such as thermal imaging~\cite{Robinette2019Sensor,Nirgudkar2022MassMIND}. 
\cite{Qiao2022Automated} reported some success of a maritime panoptic ship and buoy detection network on a private RGB dataset.
Recently the Maritime Computer Vision (MaCVi) initiative has been introduced~\cite{Kiefer20231st} with the goal of uniting the community and moving the field towards common goals. Notably, it features USV-based obstacle detection and segmentation challenges with several teams contributing approaches surpassing the previous state-of-the-art.

\paragraph{Maritime datasets.}

\begin{table}[]
    \setlength{\tabcolsep}{5pt}
    \footnotesize{
    \begin{center}
    \begin{tabular}{l@{}ccccccc}
\toprule
              & \multicolumn{1}{l}{} & \multicolumn{1}{l}{} & \multicolumn{1}{l}{}    & \multicolumn{1}{l}{} & \multicolumn{3}{c}{Classes} \\ \cmidrule{6-8} 
Dataset                                    & Frames               & T                    & Env.                  & Ann.                 & St.     & Th.     & Im.     \\ \midrule
MODD~\cite{Kristan2016Fast}                     & 4454                 & -                    & \dtSea                  & \stBox               & 1       & 2       & -       \\
MODD2~\cite{Bovcon2018Stereo}                   & 11,675               & -                    & \dtSea                  & \stBox               & 1       & 2       & -       \\
SMD~\cite{Prasad2017Video}                      & 16,000               & -                    & \dtSea                  & \stBox               & 1       & 1       & -       \\
MODS~\cite{Bovcon2020MODS}                      & 8175                 & 9                    & \dtSea                  & \stBox               & 1       & 3       & -       \\
FloW-Img~\cite{Cheng2021FloW}                   & 2000                 & -                    & \dtLake,\dtRiver        & \stBox               & -       & 1       & -       \\ \midrule
\dsUnavail{Waterline}~\cite{Steccanella2020}    & 400                  & -                    & \dtLake,\dtRiver        & \stSemantic          & 2       & -       & -       \\
Tamp-WS~\cite{Taipalmaa2019HighResolution}      & 600                  & -                    & \dtLake,\dtRiver        & \stSemantic          & 2       & -       & -       \\
USVI-WS~\cite{Cheng2021Are}                     & 700                  & -                    & \dtLake,\dtRiver        & \stSemantic          & 2       & -       & -       \\
ROSEBUD~\cite{Lambert2022ROSEBUD}               & 549                  & -                    & \dtRiver                & \stSemantic          & 7       & -       & -       \\
MaSTr1325~\cite{Bovcon2019Mastr}                & 1325                 & -                    & \dtSea                  & \stSemantic          & 4       & -       & -       \\
MaSTr1478~\cite{Zust2022Temporal}               & 1478                 & 5                    & \dtLake,\dtRiver,\dtSea & \stSemantic          & 4       & -       & -       \\ \midrule
\dsUnavail{MarPS-1395}~\cite{Qiao2022Automated} & 1395                 & -                    & \dtSea                  & \stPanoptic          & 3       & 3       & -       \\
\textbf{LaRS}                                   & 4006                 & 9                    & \dtLake,\dtRiver,\dtSea & \stPanoptic          & 3       & 8       & 19      \\ \bottomrule
    \end{tabular}
    \end{center}
    }
    \caption{Comparison of RGB-based maritime obstacle detection datasets in the number of annotated frames (Frames) and temporal context frames (T), environment types (Env.), number of stuff (St.), thing (Th.) and image-level (Im.) classes. Grayed out datasets are not publicly available. \\
    \textit{Environments}: \protect\dtLake{} - lake, \protect\dtRiver{} - river, \protect\dtSea{} - sea. \textit{Obstacle labels}: \\
    \protect\stBox{} - bounding box, \protect\stSemantic{} - semantic seg., \protect\stPanoptic{} - panoptic seg.}
    \label{tab:datasets}
\end{table}

The existing RGB maritime obstacle detection datasets are summarized in Table~\ref{tab:datasets}.
Several datasets annotate only dynamic obstacles using bounding boxes and often focus on a specific class of objects such as ships (SMD~\cite{Prasad2017Video}) or floating waste (FloW-IMG~\cite{Cheng2021FloW}).
MODD~\cite{Kristan2016Fast} and MODD2~\cite{Bovcon2018Stereo} feature more diverse dynamic obstacles annotated by bounding boxes and annotate the static obstacles by lines separating them from the water. A recent evaluation-only dataset MODS~\cite{Bovcon2020MODS} surpasses its predecessors in the number of annotated obstacles and proposes an evaluation protocol for both object detection- and segmentation-based maritime methods. The evaluation emphasizes performance aspects important for USV navigation. Two maritime datasets have been recently released in the robotics domain~\cite{DeFilippo2021RoboWhaler,Benderius2021Are}, but are not annotated for obstacle detection.

Several segmentation-oriented datasets have been proposed. A training dataset MaSTr1325~\cite{Bovcon2019Mastr} is captured in a maritime environment and annotated with per-pixel labels for water, obstacle and sky. Several smaller datasets following the same annotation protocol (Waterline~\cite{Steccanella2020}, Tampere-WaterSeg~\cite{Taipalmaa2019HighResolution} and USVInland-WS~\cite{Cheng2021Are}) were captured on inland waters, where reflections are more commonly present due to calmer waters. ROSEBUD~\cite{Lambert2022ROSEBUD} extends the number of segmentation classes, but is among the smallest datasets. 
Recently MaSTr1478~\cite{Zust2022Temporal} temporally extended~\cite{Bovcon2019Mastr} with preceding frames and included additional 153 images from inland environments featuring scenes with strong reflections. This is currently the largest maritime segmentation training dataset for obstacle detection. 
Only two panoptic maritime obstacle detection datasets have been published: MarPS-1395~\cite{Qiao2022Automated} and MassMIND~\cite{Nirgudkar2022MassMIND}. However, MarPS-1395 is not publicly available and MassMIND addresses thermal imaging only.

In short, existing public maritime datasets either lack annotations for panoptic obstacle detection or are too small for training and testing modern deep learning methods. Furthermore, they lack scene diversity since they are recorded at a single geographic location. The LaRS dataset, which we present next, overcomes these limitations and fills the gap to enable the development of the next generation of maritime obstacle detection methods.

\section{LaRS: \underline{La}kes \underline{R}ivers and \underline{S}eas dataset}

A wide range of sources was considered to ensure the visual diversity of LaRS.
Specifically, we (i) collected scenes from public online videos featuring various activities captured from boats around the world, (ii) recorded new sequences in a number of different geographic locations ourselves and (iii) included the most challenging scenes from existing maritime datasets.
 

The collection of public videos was guided using search prompts related to underrepresented scenes in the existing datasets. This includes canals (\eg "canal tour"), exotic locations (\eg "tropic boat tour", "polar kayaking"), crowded scenes (\eg "boat parade"), strong reflections (\eg "still lake"), and poor visibility conditions (\eg "boat ride in the rain", "night-time boat ride"). At least one key frame was extracted from each of the collected 396 sequences, to ensure visual diversity. In addition, a state-of-the-art obstacle segmentation network~\cite{Bovcon2021WaSR} on the collected sequences to identify additional difficult key frames.
Namely, we manually inspected the predicted segmentation and included examples with failures such as false negative obstacle segmentation and false positives on reflections to increase the difficulty level.
In this way, a set of 897 representative key frames spanning diverse and challenging scenes was selected. 

Next, we manually recorded videos at various locations on lakes, rivers and seas. From these, we identified 494 challenging sequences, and using the same process as for online videos, we identified 1354 diverse and challenging key frames.

We reviewed sequences from existing maritime datasets spanning different tasks~\cite{Prasad2017Video,Cheng2021Are,DeFilippo2021RoboWhaler} and selected 96 of the most challenging sequences -- of these, 432 key frames were selected. We also included 1323 frames from the major USV-oriented segmentation training dataset~\cite{Bovcon2019Mastr}. The collection process thus yielded a set of 4006 key frames. The contributions of individual data sources to the final set are summarized in Table~\ref{tab:frame-source}.   
 
Following~\cite{Zust2022Temporal}, to facilitate future development of detection methods that might exploit the temporal texture, we equipped all
4k key frames with the preceding 9 frames.
The total number of images in LaRS is thus over 40k. Faces were de-identified in all frames by running a face detector and blurring, followed by manual inspection.

\begin{table}
  \small{
  \begin{center}
  \begin{tabular}{lll}
    \toprule
    Source & Sequences & Key Frames \\
    \midrule
    In-house & 494 & 1354 (33.8 \%) \\
    Web videos & 396 & 897 (22.4 \%) \\
    MaSTr1325~\cite{Bovcon2019Mastr} & - & 1323 (33.0 \%) \\
    USV Inland~\cite{Cheng2021Are} & 29 & 211 (5.3 \%) \\
    MIT Sea Grant~\cite{DeFilippo2021RoboWhaler} & 35 & 122 (3.0 \%) \\
    SMD~\cite{Prasad2017Video} & 32 & 99 (2.5 \%) \\
    \bottomrule
  \end{tabular}
  \end{center}
  }
  \caption{LaRS data sources with number of the sourced sequences, the number of selected frames and their percentage in the final dataset.}
  \label{tab:frame-source}
\end{table}


\begin{figure*}
  \centering
  \includegraphics[width=\linewidth]{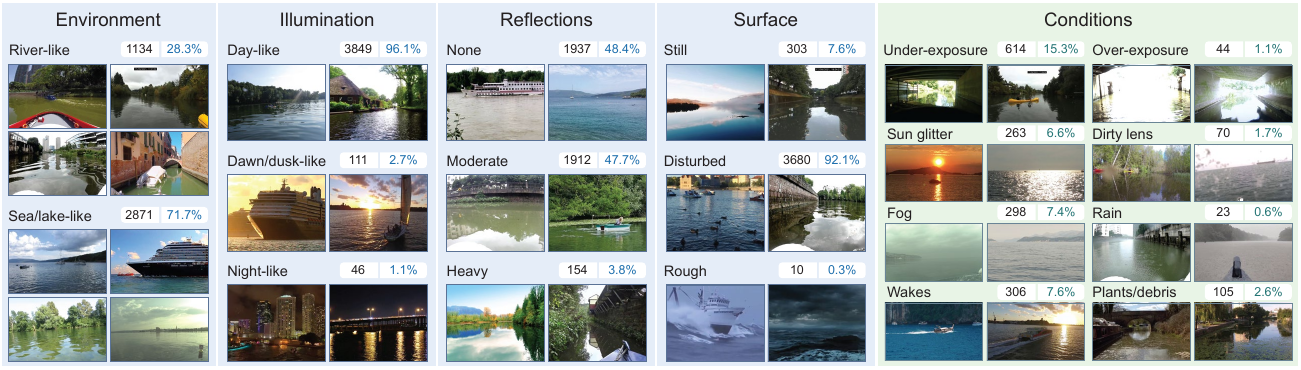}
  \caption{LaRS frames are labeled with 19 global attributes relevant for navigation.  Mutually exclusive and mutually non-exclusive groups are indicated in \textcolor{catBlue}{blue} and \textcolor{catGreen}{green}, respectively. The numbers indicate the amount of frames in the dataset.}
    \label{fig:categories}
\end{figure*}

\begin{figure}
  \centering
  \includegraphics[width=\linewidth]{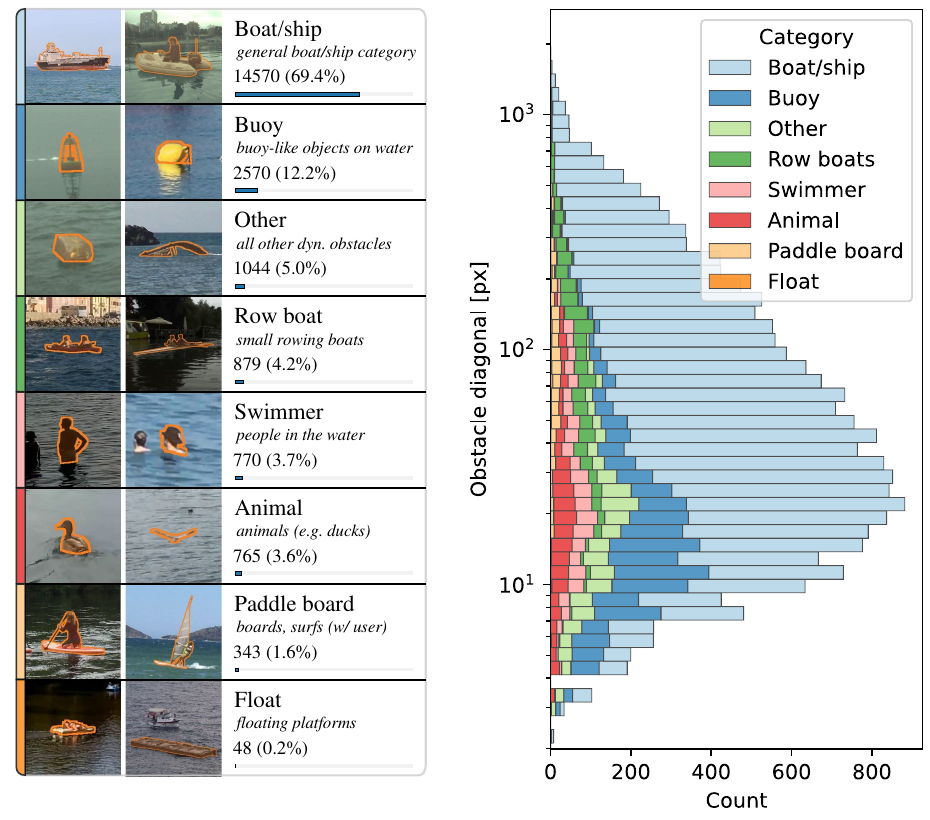}
  \caption{Statistics of dynamic obstacle classes in LaRS (left) with respect to their size (right).}
    \label{fig:size-hist}
\end{figure}

\textbf{{Dataset annotation}.} All 4k selected key frames were manually annotated with per-pixel panoptic labels by a professional labeling company.
In particular, \textit{water}, \textit{sky} and \textit{static obstacles} like shores and piers were annotated as stuff classes, while the dynamic obstacles instances were segmented and classified into 8 different object categories (see Figure~\ref{fig:size-hist}): \textit{boat}, \textit{row boat}, \textit{paddle board}, \textit{buoy}, \textit{swimmer}, \textit{animal}, \textit{float} and an open-world \textit{other} class to cover the remaining obstacles. Following a standard practice~\cite{Lin2014COCO} \textit{group labels} were used to group multiple hard-to-delineate neighbouring instances of the same category. Regions that could not be reliably manually segmented were labeled with the \textit{ignore} class. Global attributes were assigned to key frames, to indicate \textit{environment type}, \textit{illumination conditions}, \textit{presence of reflections}, \textit{surface roughness} and \textit{scene conditions}. Examples of scenes corresponding to the 19 global attribute labels are shown in Figure~\ref{fig:categories}.

Annotation correctness was further analyzed to ensure the highest quality of the dataset. In the first pass, state-of-the-art semantic segmentation and panoptic segmentatation methods were trained and run on the entire dataset to identify major annotation errors. Visual inspection of large FP and FN predictions revealed annotation errors in 210 images, which were manually corrected. Finally, we manually inspected all ground truth instance labels of the dynamic obstacles and identified and corrected approximately 3600 annotation errors. The statistics of the final dynamic obstacle categories their instance distribution by size are shown in Figure~\ref{fig:size-hist}.

\textbf{Dataset evaluation splits.} The dataset was split into training (65 \%), validation (5 \%) and test (30 \%) sets. To prevent overfitting, we made sure there was no overlap between the sets, i.e., that all key frames extracted from a single sequence are contained within the same set. 
We also ensured that the distribution of the resolution, reflection levels and scene types is similar across the dataset splits. This was done by computing histograms over the aforementioned properties within each set and computing the Hellinger distances between all three pairs of image sets. A randomized search was then applied to create splits that minimized the average Hellinger distance. 
The training and test splits will be publicly released along with the ground truth. For the test set, only the frames will be released, while the ground truth is withheld and an evaluation server has been set-up to provide automated and unbiased evaluation.

\section{Evaluation protocol}\label{sec:protocol}

The methods are trained on the training set, the validation set is used for stopping criterion and the performance is evaluated on the test set. The evaluation protocol includes two tasks: (i) the classical semantic-segmentation-based obstacle detection and (ii) panoptic-segmentation-based obstacle detection. The respective performance measures are described next. 



\subsection{Semantic segmentation performance measures}\label{sec:protocol/semantic}

The standard maritime obstacle detection evaluation protocol MODS~\cite{Bovcon2020MODS} is applied to analyze the methods based on semantic segmentation. This protocol considers three semantic classes: water, sky and obstacle. The first two are directly obtained from the ground truth panoptic labels, while the last is obtained by combining all dynamic and static obstacle annotations. In addition to MODS domain-specific primary measures, we also compute the mean intersection-over-union (mIoU), a commonly used measure in general semantic segmentation~\cite{Lin2014COCO,Cordts2016Cityscapes,Geiger2012Are}.

The MODS primary performance measures are (i) water-edge estimation accuracy computed from boundary between water and static obstacles and (ii) dynamic obstacle detection accuracy. The ground truth panoptic labels simplify the water-edge estimation accuracy measure, which we define as per-pixel classification accuracy evaluated within a $d$ pixels thick region around the ground-truth water edge, $G_d$, i.e.,
\begin{equation}
    \mu = \frac{1}{|G_d|}\sum\nolimits_{(p,g) \in G_d} [p = g],
\end{equation}
where $p$ and $g$ are predicted and ground-truth labels of pixels in $G_d$.

The MODS dynamic obstacle detection accuracy is determined by precision (Pr), recall (Re) and F1 score calculated in correspondence to the practical use of the methods. The method iterates over all ground truth dynamic obstacles. If the coverage of the predicted \textit{obstacle} pixels exceeds $\theta = 0.7$, the dynamic obstacle is counted as a true-positive, otherwise it counts as a false-negative. The number of false-positives is estimated as the number of predicted obstacle segments (computed by connected components) in the ground-truth water mask. Please see~\cite{Bovcon2020MODS} for further details.

\subsection{Panoptic segmentation performance measures}

Standard panoptic performance evaluation measures~\cite{Kirillov2019Panoptic} are used: segmentation quality (SQ), recognition quality (RQ) and the combined panoptic quality (PQ):
%
\begin{equation}
     \text{PQ} = \underbrace{\frac{\sum_{(p,g) \in \text{TP}} \text{IoU}(p, g)}{|\text{TP}|}}_\text{segmentation quality (SQ)} \times \underbrace{\frac{|\text{TP}|}{|\text{TP}| + \frac{1}{2}|\text{FP}| + \frac{1}{2}|\text{FN}|}}_\text{recognition quality (RQ)}.
\end{equation}
%
The individual metrics are also reported separately for \textit{thing} and \textit{stuff} classes indicated by superscripts $(\cdot)^\text{Th}$ and $(\cdot)^\text{St}$. 
 

It should be noted that, from the perspective of obstacle detection, additional instance detections on static obstacles are not considered false positives. Additionally, misclassification of an obstacle type is considered less critical than failing to detect the obstacle altogether. Therefore, we also report obstacle-class-agnostic variants of the metrics, which ignore the class label, denoted by $(\cdot)^{\text{Th}_a}$.

\section{Experimental results}

\subsection{Semantic segmentation methods}\label{sec:exp/semantic}

We considered 19 methods. Three single-frame state-of-the-art maritime-specific obstacle detection methods (WaSR~\cite{Bovcon2021WaSR}, WODIS~\cite{Chen2021WODIS}, IntCatchAI~\cite{Steccanella2020}) and several general semantic segmentation methods, i.e., four
FNC-style classical methods (FCN~\cite{Long2015Fully}, UNet~\cite{Ronneberger2015UNet}, DeepLabv3~\cite{Chen2017Rethinking}, DeepLabv3+~\cite{Chen2018Encoder}, PointRend~\cite{Kirillov2020PointRend}, KNet~\cite{Zhang2021KNet}), three modern lightweight convolutional methods (BiSeNetv1~\cite{Yu2018Bisenet}, BiSeNetv2~\cite{Yu2021BiSeNet}, STDC~\cite{Fan2021Rethinking}) and two transformer-based methods (SegFormer~\cite{Xie2021SegFormer}, Segmenter~\cite{Strudel2021Segmenter}). The selection also includes two recent temporal semantic segmentation methods from the AGV domain (CSANet~\cite{Yuan2021CSANet}, TMANet~\cite{Wang2021Temporal}) and one from maritime domain (WaSR-T~\cite{Zust2022Temporal}).

\begin{table}
    \setlength{\tabcolsep}{3pt}
    {\scriptsize
    \begin{center}
    \begin{tabular}{llccccccc}
    \toprule
    Architecture & Bbone &  $\mu$ &    Pr &    Re &    F1 &  mIoU & FPS & GMacs \\
    \midrule
       UNet~\cite{Ronneberger2015UNet} &         S5 &       75.7 &        8.6 &       70.6 &       15.4 &       90.1 &        5.2 &     1621 \\
              FCN~\cite{Long2015Fully} &  RN-50 &       76.8 &       50.1 &       68.7 &       57.9 &       92.6 &        5.2 &     1582 \\
              FCN~\cite{Long2015Fully} & RN-101 &       77.4 &       59.0 &       68.5 &       63.4 &       95.0 &        3.4 &     2203 \\
   DeepLabv3~\cite{Chen2017Rethinking} & RN-101 &       77.5 &       61.1 &       72.0 & \bm3{66.1} &       95.2 &        2.4 &     2779 \\
     DeepLabv3+~\cite{Chen2018Encoder} & RN-101 & \bm3{77.8} &       57.8 &       71.7 &       64.0 &       95.4 &        3.3 &     2031 \\
PointRend~\cite{Kirillov2020PointRend} & RN-101 &       77.5 &       60.6 &       71.1 &       65.4 &       94.9 &        8.7 &      521 \\
        BiSeNetv1~\cite{Yu2018Bisenet} &  RN-50 &       73.3 &       31.6 &       66.3 &       42.8 &       92.2 &       10.1 &      792 \\
        BiSeNetv2~\cite{Yu2021BiSeNet} &          - &       73.9 &       48.2 &       63.2 &       54.7 &       93.5 & \bm3{51.1} &       98.4 \\
        STDC1~\cite{Fan2021Rethinking} &          - &       75.6 &       58.6 &       65.3 &       61.8 &       93.6 & \bm1{72.9} & \bm3{67.7} \\
        STDC2~\cite{Fan2021Rethinking} &          - &       76.5 & \bm2{64.3} &       64.3 &       64.3 &       94.5 & \bm2{56.5} &       94.1 \\
     SegFormer~\cite{Xie2021SegFormer} &     MiT-B2 & \bm2{78.6} & \bm3{63.8} & \bm2{77.5} & \bm2{70.0} & \bm2{96.8} &        5.6 &      144 \\
 Segmenter~\cite{Strudel2021Segmenter} &      ViT-B &       72.2 &       51.6 &       59.5 &       55.2 &       95.1 &        2.6 &      556 \\
             KNet~\cite{Zhang2021KNet} &     Swin-T & \bm1{78.8} & \bm1{67.6} & \bm1{80.4} & \bm1{73.4} & \bm1{97.2} &        4.2 &     1973 \\
    \midrule
            WaSR~\cite{Bovcon2021WaSR} & RN-101 &       71.0 &       59.9 &       63.4 &       61.6 &       96.6 &       16.5 &      399 \\
            WODIS~\cite{Chen2021WODIS} & RN-101 &       63.0 &       38.8 &       61.1 &       47.5 &       85.7 &       35.4 & \bm2{61.8} \\
     IntCatchAI~\cite{Steccanella2020} &          - &       62.4 &       40.6 &       50.2 &       44.9 &       45.6 &        6.7 &  \bm1{4.7} \\
    \midrule
        WaSR-T~\cite{Zust2022Temporal} & RN-101 &       71.1 &       59.7 &       64.7 &       62.1 & \bm3{96.7} &       11.1 &      579 \\
          CSANet~\cite{Yuan2021CSANet} & RN-101 &       63.7 &       47.2 &       58.2 &       52.1 &       94.2 &        2.1 &     3912 \\
        TMANet~\cite{Wang2021Temporal} &  RN-50 &       77.1 &       52.5 & \bm3{73.4} &       61.2 &       94.1 &        0.8 &     5193 \\
    \bottomrule
    \end{tabular}
    \end{center}
    }
    \caption{Performance of single-image state-of-the-art general (top), maritime (middle) and temporal (bottom) semantic segmentation methods on LaRS. Gold, silver and bronze indicate the top three scores in each category.}
    \label{tab:semantic}
\end{table}

\begin{figure}
  \centering
  \includegraphics[width=\linewidth]{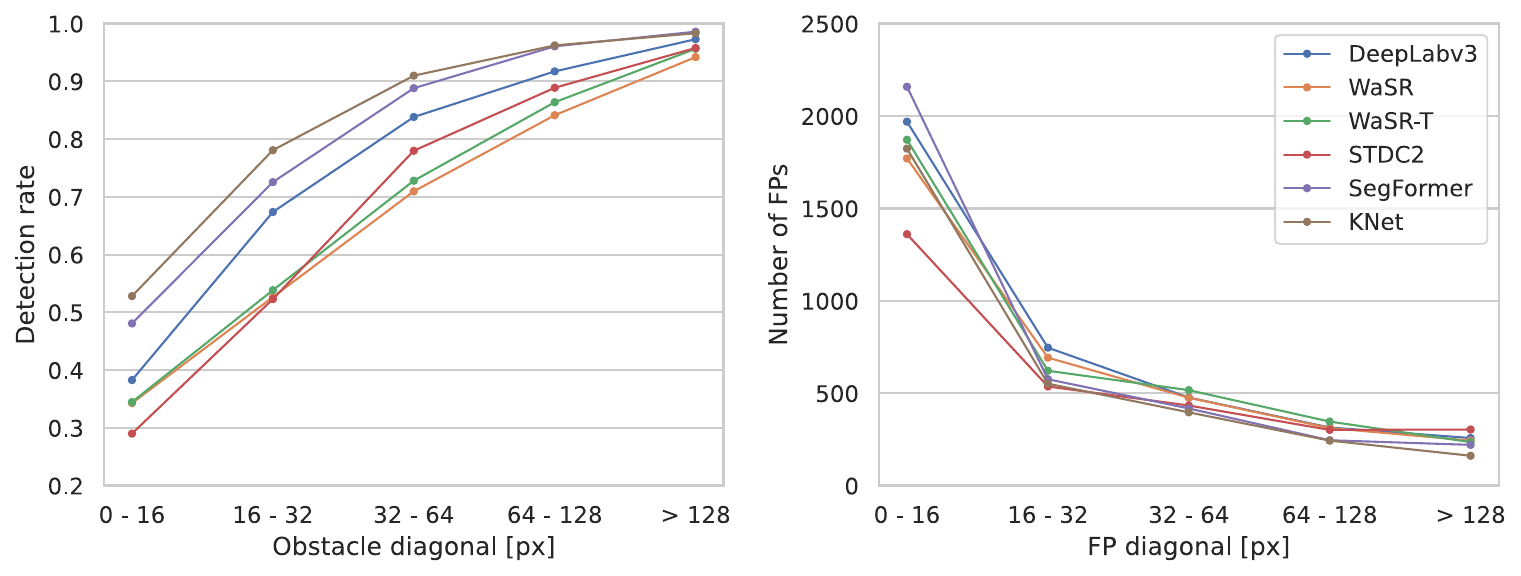}
  \caption{Segmentation-based obstacle detection rate (left) and number of false positives (right) w.r.t. the obstacle size.}
    \label{fig:semantic-size-re}
\end{figure} 

\begin{figure*}
  \centering
  \includegraphics[width=\linewidth]{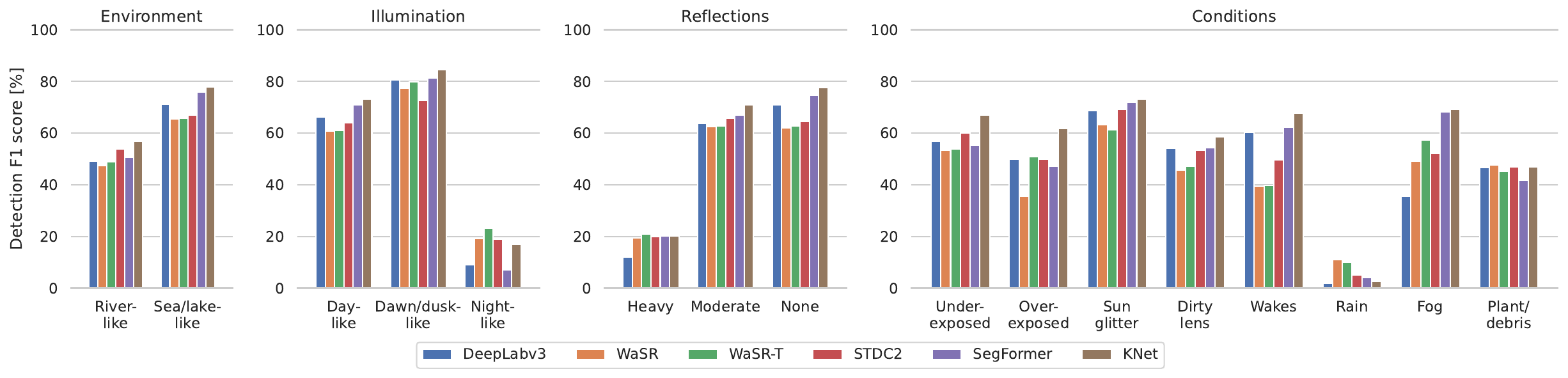}
  \caption{Semantic segmentation detection performance (F1) with respect to global attributes.}
    \label{fig:semantic-cat}
\end{figure*} 

\begin{figure*}
  \centering
  \includegraphics[width=\linewidth]{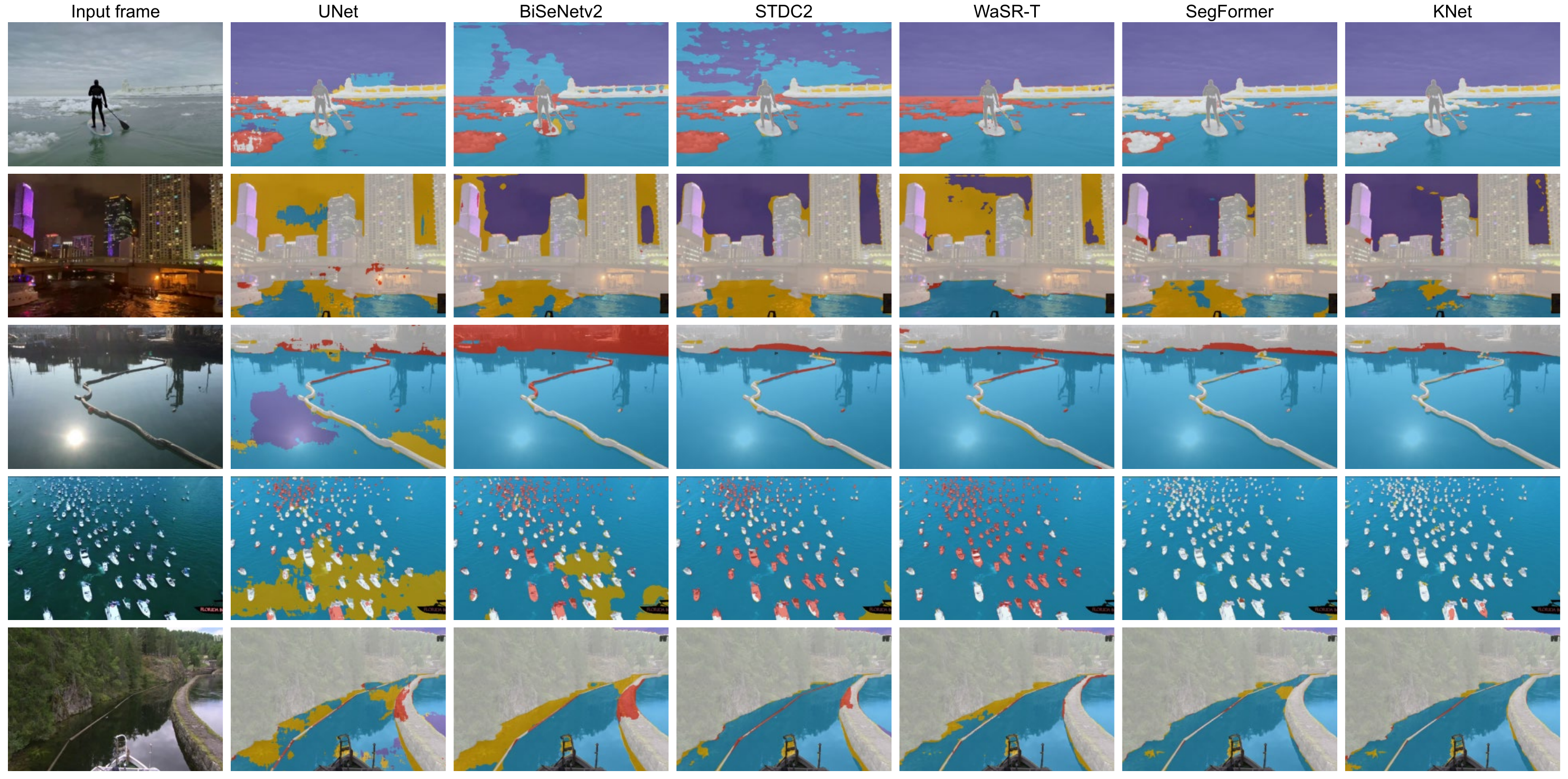}
  \caption{Qualitative semantic segmentation results on LaRS. Sky and water classes are shown in purple and blue, respectively. TP, FN and FP obstacle predictions are shown in white, red and yellow, respectively, while black indicates the ignore region.}
    \label{fig:segmentation-qualitative}
\end{figure*}


WaSR~\cite{Bovcon2021WaSR}, WaSR-T~\cite{Zust2022Temporal}, CSANet~\cite{Yuan2021CSANet}, TMANet~\cite{Wang2021Temporal}, WODIS~\cite{Chen2021WODIS} and IntCatchAI~\cite{Steccanella2020} were trained using their official configurations. 
All other methods were trained using \textit{MMSegmentation}~\cite{mmseg2020} with their Cityscapes configurations adapted to LaRS. The methods were trained on 2 x NVIDIA V100 GPUs with a batch size of 8.
Runtimes were estimated in frames per second (FPS) on a single GPU.


The results are reported in Table~\ref{tab:semantic}. KNet~\cite{Zhang2021KNet} achieves the best  water-edge accuracy (78.8 \%), followed by SegFormer~\cite{Xie2021SegFormer} (-0.2\%), which implies a very good segmentation accuracy. 
This is supported by mIoU, which ranks these two methods at the top. More importantly, these two methods also outperform all other methods in F1 score by a large margin, indicating very good dynamic obstacle detection performance. Specifically, KNet ranks first, followed by SegFormer (-3.4\% F1 score) and DeepLabv3~\cite{Chen2017Rethinking} (-7.3\% F1 score).
 
 
Note that the best-performing methods are relatively slow ({\textapp} 4-5 FPS) even on high-end hardware and may not be suitable for real-world applications with often limited compute power. Alternatively, STDC1 and STDC2~\cite{Fan2021Rethinking} demonstrate exceptional efficiency ({\textapp} 50-70 FPS), while incurring a performance drop of 9-10\% in terms of F1 score compared to the top performer KNet.


\begin{table*}
    \setlength{\tabcolsep}{3pt}
    {\footnotesize
    \begin{center}
    \begin{tabular}{llccccccccccccccccc}
    \toprule
     &  & \multicolumn{4}{c}{PQ (\%)} &  & \multicolumn{4}{c}{RQ (\%)} & & \multicolumn{4}{c}{SQ (\%)} & & \\
     \cmidrule{3-6} \cmidrule{8-11} \cmidrule{13-16}
     Architecture & Backbone &    All &    Th &    Th$_a$ &  St  &  & All &    Th &    Th$_a$ &  St & & All &    Th &    Th$_a$ &  St & & FPS & GMacs\\
    \midrule
Panoptic Deeplab~\cite{Cheng2020Panoptic} &  ResNet-50 &       34.7 &       13.4 &       33.0 &       91.4 &    &       40.3 &       19.3 & \bm1{46.3} &       96.2 &    &       69.5 &       60.0 &       71.3 &       94.9 &    &        6.0 & \bm1{339.3} \\
Panoptic FPN~\cite{Kirillov2019Panoptica} &  ResNet-50 & \bm2{40.1} & \bm2{21.7} & \bm1{35.5} &       89.3 &    & \bm2{46.9} & \bm2{28.6} & \bm3{45.9} &       95.8 &    & \bm3{73.5} & \bm2{66.1} & \bm1{77.3} &       93.1 &    & \bm1{21.7} &       471.4 \\
Panoptic FPN~\cite{Kirillov2019Panoptica} & ResNet-101 &       38.7 & \bm3{19.7} & \bm1{35.5} &       89.4 &    &       45.0 & \bm3{26.1} & \bm2{46.0} &       95.5 &    & \bm2{73.6} & \bm2{66.1} & \bm2{77.1} &       93.5 &    & \bm2{16.7} &       627.2 \\
    MaX-DeepLab~\cite{Wang2020MaXDeepLab} &      MaX-S &       31.9 &        9.5 &       19.2 &       91.7 &    &       36.1 &       13.4 &       26.0 &       96.6 &    &       71.3 &       62.5 &       73.7 &       94.8 &    &        3.7 &           - \\
  Mask2Former~\cite{Cheng2021Mask2Former} &  ResNet-50 &       37.6 &       17.0 &       27.9 &       92.4 &    &       43.7 &       23.6 &       37.6 & \bm3{97.3} &    &       71.3 &       62.4 &       74.2 &       95.0 &    & \bm3{10.6} & \bm2{464.2} \\
  Mask2Former~\cite{Cheng2021Mask2Former} & ResNet-101 &       37.2 &       16.3 &       29.2 & \bm3{92.8} &    &       43.0 &       22.7 &       38.9 &       97.1 &    &       71.4 &       62.3 &       75.0 & \bm2{95.5} &    &        5.7 &       620.0 \\
  Mask2Former~\cite{Cheng2021Mask2Former} &     Swin-T & \bm3{39.2} &       18.8 & \bm2{34.0} & \bm2{93.7} &    & \bm3{45.5} &       25.8 &       45.2 & \bm2{98.1} &    &       72.2 & \bm3{63.5} &       75.2 & \bm3{95.4} &    &        5.4 & \bm3{470.7} \\
  Mask2Former~\cite{Cheng2021Mask2Former} &     Swin-B & \bm1{41.7} & \bm1{21.8} & \bm3{33.6} & \bm1{94.7} &    & \bm1{48.5} & \bm1{29.7} &       44.6 & \bm1{98.5} &    & \bm1{78.2} & \bm1{71.5} & \bm3{75.3} & \bm1{96.2} &    &        4.8 &       948.0 \\
    \bottomrule
    \end{tabular}
    \end{center}
    }
    \caption{Panoptic quality (PQ), recognition quality (RQ) and segmentation quality (SQ) reported overall (All) and with respect to stuff (St) and things (Th), with Th$_a$ denoting class-agnostic score. The inference speed is reported in FPS.}
    \label{tab:panoptic-normal}
\end{table*}

\begin{table}
    {\small
    \begin{center}
    \begin{tabular}{llccc}
    \toprule
     Architecture & Backbone &    $\mu$ &    F1 &    mIoU \\
    \midrule
Panoptic Deeplab~\cite{Cheng2020Panoptic} &  ResNet-50 &       73.5 & \bm2{64.6} &       95.4 \\
Panoptic FPN~\cite{Kirillov2019Panoptica} &  ResNet-50 &       67.2 &       58.9 &       93.5 \\
Panoptic FPN~\cite{Kirillov2019Panoptica} & ResNet-101 &       66.9 &       58.1 &       93.3 \\
    MaX-DeepLab~\cite{Wang2020MaXDeepLab} &      MaX-S &       72.7 & \bm3{60.2} &       95.4 \\
  Mask2Former~\cite{Cheng2021Mask2Former} &  ResNet-50 &       75.1 &       54.9 &       95.4 \\
  Mask2Former~\cite{Cheng2021Mask2Former} & ResNet-101 & \bm3{75.8} &       53.2 & \bm3{95.6} \\
  Mask2Former~\cite{Cheng2021Mask2Former} &     Swin-T & \bm2{76.2} &       56.7 & \bm2{96.8} \\
  Mask2Former~\cite{Cheng2021Mask2Former} &     Swin-B & \bm1{77.4} & \bm1{71.1} & \bm1{97.6} \\
    \bottomrule
    \end{tabular}
    \end{center}
    }
    \caption{Performance of panoptic methods under the semantic segmentation setup.}
    \label{tab:panoptic-semantic}
\end{table}

To further probe the performance of the best-performing and fastest methods, we analyze the detection rate (Re) and the number of FP detections with respect to the obstacle size in Figure~\ref{fig:semantic-size-re}.
The largest performance variance between methods is observed for small obstacles. This is where KNet and SegFormer most substantially stand out from the rest, which is also confirmed by qualitative examples in Figure~\ref{fig:segmentation-qualitative}, particularly on thin (third row) and compact small obstacles (fourth row).

Interestingly, compared to single-frame methods, the temporal methods do not appear to benefit from the additional temporal context. For example, the performance of temporal WaSR-T~\cite{Zust2022Temporal} is almost on par (+0.5\% F1) with its single-frame counterpart WaSR~\cite{Bovcon2021WaSR}. Since the prior work~\cite{Zust2022Temporal} on a smaller training set indicated a clear advantage of WaSR-T over WaSR, we speculate that the observed reduced difference is due to the increased size and larger diversity of the LaRS training set.

Figure~\ref{fig:semantic-cat} investigates performance with respect to scene attributes. Overall, river-like environments are more challenging compared to sea/lake-like environments, which may be attributed to a larger quantity of reflections and background variety of the former. The methods are fairly robust to dusk scenes, with a moderate performance increase compared to daytime scenes. However, the performance substantially drops on night-time scenes. Interestingly, a performance advantage of the temporal WaSR-T is observed over the single-frame counterparts, which indicates the potential for exploiting temporal context in situations with significant visual ambiguity. Moreover, all methods are fairly robust to moderate reflections, while strong reflections lead to substantial performance drops. Of the different scene conditions, the methods perform best on sun glitter, fog, and wakes, while the worst performance is observed in the presence of rain, dirty lenses, and plants/debris.

\subsection{Panoptic segmentation methods}

\begin{figure*}
  \centering
  \includegraphics[width=\linewidth]{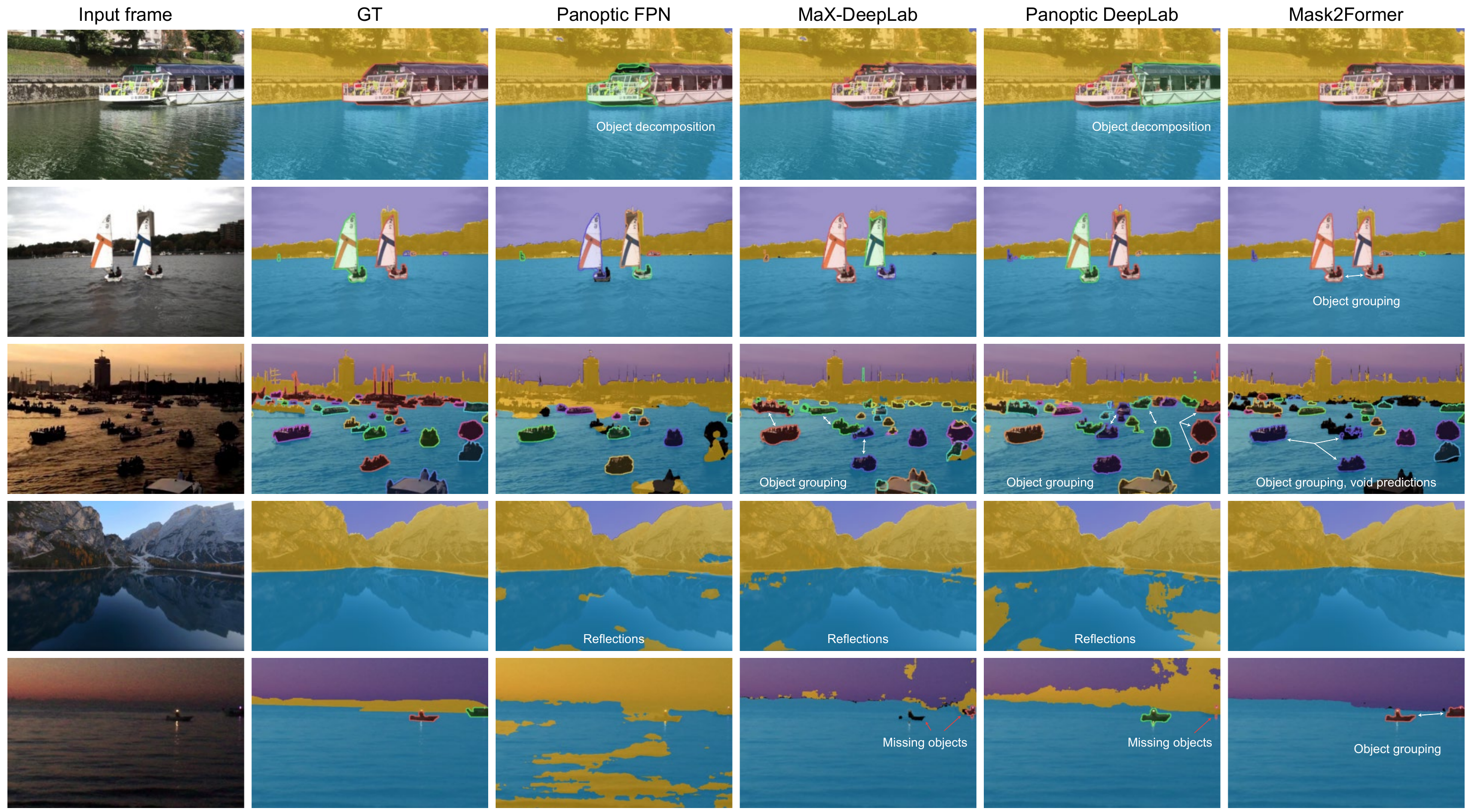}
  \caption{Qualitative panoptic segmentation results. Individual instance detections are outlined with different colors. Void predictions are colored black. Common errors are indicated with white text.}
    \label{fig:panoptic-qualitative}
\end{figure*}

\begin{figure}
  \centering
  \includegraphics[width=\linewidth]{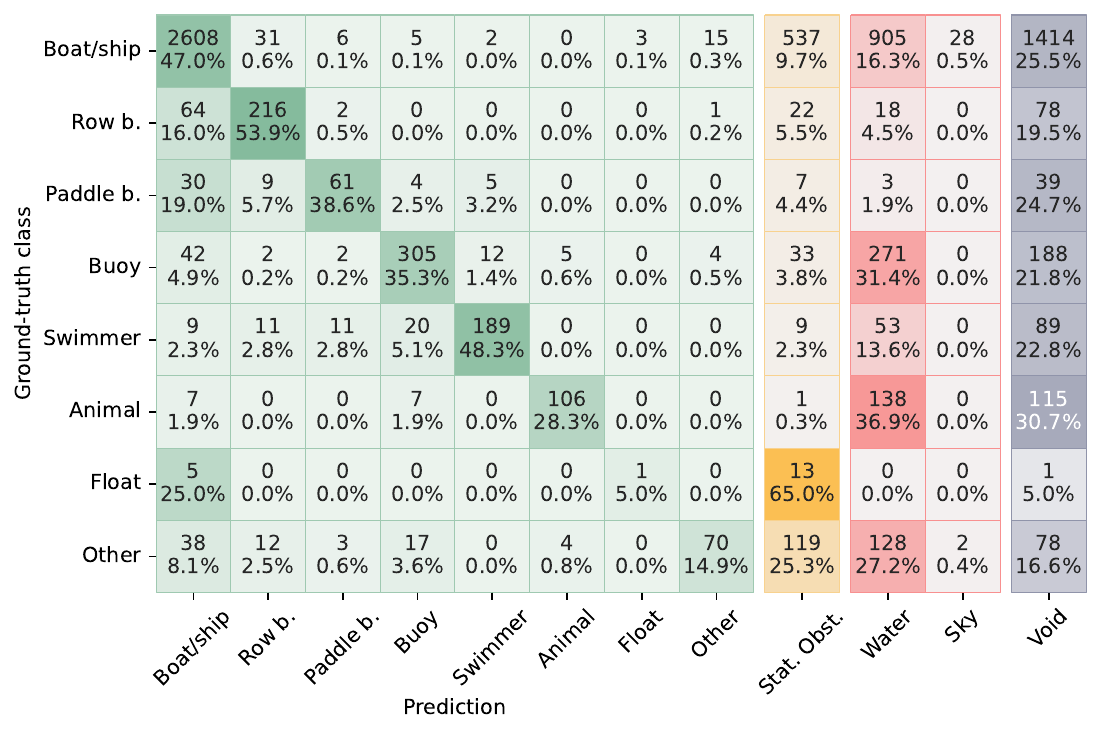}
  \caption{Confusion matrix of ground-truth dynamic obstacles for Mask2Former with the Swin-B backbone.}
    \label{fig:conf-matrix}
\end{figure}


Several panoptic methods with various backbones are considered: Panoptic Deeplab~\cite{Cheng2020Panoptic} and Panoptic FPN~\cite{Kirillov2019Panoptica} as members of conv-net family with strong baseline performance on ground-vehicle-related tasks, and two state-of-the-art representatives of transformer-based mask classification methods MaX-Deeplab~\cite{Wang2020MaXDeepLab} and Mask2Former~\cite{Cheng2021Mask2Former}. The methods were trained on 2 x NVIDIA V100 GPUs with a batch size of 4. 


Results in Table~\ref{tab:panoptic-normal} indicate that the top PQ performance is achieved by Swin-B-based Mask2Former~\cite{Cheng2021Mask2Former} (41.7 \%), followed by Panoptic FPN~\cite{Kirillov2019Panoptica} (-1.6 \%) and Swin-T-based Mask2Former (-2.5 \%). Overall, the methods achieve relatively low PQ scores. Comparing PQ$^{\text{Th}}$ and PQ$^{\text{St}}$, we observe that the static obstacles (i.e., stuff class) are well detected (PQ$^{\text{St}}=94.7\%$ for the best method) but methods struggle the detection of dynamic obstacles (i.e., things). 

Specifically, the recognition quality for dynamic obstacles of the best method is only RQ$^{\text{Th}}=27.7 \%$. High RQ$^{\text{Th}}$ requires accurate detection obstacles as well as correct classification. Ignoring the classification errors (RQ$^{\text{Th}_a}$) substantially increases this score (to 44.6 \%), which confirms that a major source of errors is obstacle misclassification. 
We thus plot the confusion matrix between predicted and GT instance classes for the top performing method (Swin-B-based Mask2Former~\cite{Cheng2021Mask2Former}) in Figure~\ref{fig:conf-matrix} and observe significant confusion between boat/ship, row boats, paddle board and float categories. The objects from the rarer classes are often predicted as the more common \mbox{boat/ship} category. In addition, similarly to what we observed in semantic segmentation methods, small obstacles such as buoys, swimmers and animals are often missed and segmented as water. 
It should be noted that modern panoptic methods use a \textit{void} label for regions without sufficiently confident segment predictions. Void labels account for approximately 24\% of all predictions on dynamic obstacles.

Another source of errors is the grouping of objects into a single detection and the decomposition of a single instance into several detections. 
The qualitative examples in Figure~\ref{fig:panoptic-qualitative} show that incorrect object grouping/splitting is particularly acute in dense scenes (row 3).
Interestingly, the best-performing method Mask2Former sometimes incorrectly groups even well-separated instances (rows 2 and 3).

Note that labeling several small water regions as static obstacles substantially affects robotic navigation in practice, since the USV might frequently stop to avoid a possible collision. This is not properly reflected in panoptic performance measures, which would decrease only slightly.
Similarly, joining two nearby obstacles into a single instance is not detrimental from a practical obstacle avoidance standpoint, but can significantly reduce the panoptic measures.

We thus also evaluate the methods with semantic segmentation measures from Section~\ref{sec:protocol/semantic}, by assigning all detected static and dynamic obstacles and void predictions to the \textit{obstacle} class. Results in Table~\ref{tab:panoptic-semantic} reveal that the best panoptic methods perform on par with state-of-the-art semantic segmentation methods under this setup. For example, the best panoptic method (Mask2Former with Swin-B backbone) lags behind the best semantic segmentation method (Table~\ref{tab:semantic}) by only -2.3 \% in F1 score. This presents a clear opportunity for panoptic methods, whose performance would greatly improve also at the panoptic level by properly addressing the instance detection and separation capability.

\subsection{Difficulty level of LaRS}

We conduct an experiment to demonstrate the difficulty level of the LaRS benchmark. We train the best performing semantic segmentation method KNet on some of the largest and most diverse existing maritime segmentation datasets MaSTr1325~\cite{Bovcon2019Mastr}, MaSTr1478~\cite{Zust2022Temporal} and ROSEBUD~\cite{Lambert2022ROSEBUD} and evaluate them on the LaRS test set. Results are presented in Table~\ref{tab:diversity}.

We observe a severe performance drop when training on previously available datasets. These datasets are limited in nature and lack the variety required to tackle the LaRS benchmark. For example MaSTr1325 only contains maritime scenes, while ROSEBUD only contains fluvial scenes. Furthermore, even combining all the examples from related datasets for training the network is not enough and leads to subpar performance compared to LaRS training (F1 drop of 12.8 \%). This suggest that the current datasets are just not representative enough for general maritime perception and outlines the need for large, diverse datasets like LaRS to move the field forward.

\begin{table}
    {\footnotesize
    \begin{center}
    \begin{tabular}{lccccc}
    \toprule
    Train dataset &  $\mu$ &    Pr &    Re &    F1 &  mIoU \\
    \midrule
          MaSTr1325 &    62.2 & 28.2 & 69.9 & 40.2 &  87.5 \\
          MaSTr1478 &    72.5 & 52.1 & 67.0 & 58.6 &  93.6 \\
            ROSEBUD &    64.5 & 30.1 & 57.2 & 39.5 &  81.6 \\
MaSTr1478 + ROSEBUD &    72.2 & 55.0 & 67.5 & 60.6 &  92.9 \\
    \midrule
               LaRS &    \textbf{78.8} & \textbf{67.6} & \textbf{80.4} & \textbf{73.4} &  \textbf{97.2} \\
    \bottomrule
    \end{tabular}
    \end{center}
    }
    \caption{Performance of KNet semantic segmentation on the LaRS test set, when trained with different existing maritime segmentation datasets.}
    \label{tab:diversity}
\end{table}

\section{Conclusion}

We presented the first maritime panoptic obstacle detection benchmark LaRS, containing scenes from \underline{la}kes, \underline{r}ivers, and \underline{s}eas. LaRS is the largest dataset of its kind and exceeds other maritime obstacle detection datasets in terms of the diversity of recording locations, acquisition conditions, obstacle appearances, number of categories and annotation detail. Each key frame is annotated by panoptic segmentation labels, 19 global attributes and additionally equipped with several preceding frames to enable the development of methods exploiting temporal context. 


Results for 27 semantic- and panoptic-segmentation-based detection methods reveal that semantic-segmentation methods slightly outperform the panoptic counterparts in overall segmentation quality. We identify several opportunities for improvement of the methods, notably improving the instance separation of panoptic methods and better exploitation of the temporal context in scenes with significant ambiguity.
The dataset, tookit and the online evaluation server will be publicly released to foster further advancements in maritime obstacle detection.


\section*{Acknowledgments}
This work was supported by the Slovenian Research Agency programs P2-0214 and P2-0095, and project J2-2506.

{\small
\bibliographystyle{ieee_fullname}
\bibliography{references}

\begin{thebibliography}{10}\itemsep=-1pt

\bibitem{Benderius2021Are}
Ola Benderius, Christian Berger, and Krister Blanch.
\newblock Are we ready for beyond-application high-volume data? {{The Reeds}}
  robot perception benchmark dataset, Sept. 2021.

\bibitem{Bovcon2021WaSR}
Borja Bovcon and Matej Kristan.
\newblock {{WaSR--A Water Segmentation}} and {{Refinement Maritime Obstacle
  Detection Network}}.
\newblock {\em IEEE Transactions on Cybernetics}, pages 1--14, July 2021.

\bibitem{Bovcon2018Stereo}
Borja Bovcon, Rok Mandeljc, Janez Per{\v s}, and Matej Kristan.
\newblock Stereo obstacle detection for unmanned surface vehicles by
  {{IMU-assisted}} semantic segmentation.
\newblock {\em Robotics and Autonomous Systems}, 104, 2018.

\bibitem{Bovcon2019Mastr}
Borja Bovcon, Jon Muhovi{\v c}, Janez Per{\v s}, and Matej Kristan.
\newblock The {{MaSTr1325}} dataset for training deep {{USV}} obstacle
  detection models.
\newblock In {\em 2019 {{IEEE}}/{{RSJ International Conference}} on
  {{Intelligent Robots}} and {{Systems}} ({{IROS}})}, pages 3431--3438, 2019.

\bibitem{Bovcon2020MODS}
Borja Bovcon, Jon Muhovi{\v c}, Du{\v s}ko Vranac, Dean Mozeti{\v c}, Janez
  Per{\v s}, and Matej Kristan.
\newblock {{MODS}} -- {{A USV-oriented}} object detection and obstacle
  segmentation benchmark.
\newblock {\em IEEE Transactions on Intelligent Transportation Systems}, May
  2021.

\bibitem{Cane2016SaliencyBased}
Tom Cane and James Ferryman.
\newblock Saliency-{{Based Detection}} for {{Maritime Object Tracking}}.
\newblock In {\em 2016 {{IEEE Conference}} on {{Computer Vision}} and {{Pattern
  Recognition Workshops}} ({{CVPRW}})}, pages 1257--1264, June 2016.

\bibitem{Cane2019Evaluating}
Tom Cane and James Ferryman.
\newblock Evaluating deep semantic segmentation networks for object detection
  in maritime surveillance.
\newblock In {\em Proceedings of {{AVSS}} 2018 - 2018 15th {{IEEE International
  Conference}} on {{Advanced Video}} and {{Signal-Based Surveillance}}}, 2019.

\bibitem{Carion2020EndtoEnd}
Nicolas Carion, Francisco Massa, Gabriel Synnaeve, Nicolas Usunier, Alexander
  Kirillov, and Sergey Zagoruyko.
\newblock End-to-{{End Object Detection}} with {{Transformers}}.
\newblock {\em Lecture Notes in Computer Science (including subseries Lecture
  Notes in Artificial Intelligence and Lecture Notes in Bioinformatics)}, 12346
  LNCS:213--229, May 2020.

\bibitem{Chan2021SegmentMeIfYouCan}
Robin Chan, Krzysztof Lis, Svenja Uhlemeyer, Hermann Blum, Sina Honari, Roland
  Siegwart, Mathieu Salzmann, Pascal Fua, and Matthias Rottmann.
\newblock {{SegmentMeIfYouCan}}: {{A Benchmark}} for {{Anomaly Segmentation}}.
\newblock Apr. 2021.

\bibitem{Chen2017Rethinking}
Liang~Chieh Chen, George Papandreou, Florian Schroff, and Hartwig Adam.
\newblock Rethinking atrous convolution for semantic image segmentation.
\newblock {\em arXiv preprint arXiv:1706.05587}, June 2017.

\bibitem{Chen2018Encoder}
Liang-Chieh Chen, Yukun Zhu, George Papandreou, Florian Schroff, and Hartwig
  Adam.
\newblock Encoder-{{Decoder}} with {{Atrous Separable Convolution}} for
  {{Semantic Image Segmentation}}.
\newblock In {\em Proceedings of the {{European}} Conference on Computer Vision
  ({{ECCV}})}, pages 801--818, Feb. 2018.

\bibitem{Chen2021WODIS}
Xiang Chen, Yuanchang Liu, and Kamalasudhan Achuthan.
\newblock {{WODIS}}: {{Water Obstacle Detection Network Based}} on {{Image
  Segmentation}} for {{Autonomous Surface Vehicles}} in {{Maritime
  Environments}}.
\newblock {\em IEEE Transactions on Instrumentation and Measurement}, 70:1--13,
  2021.

\bibitem{Cheng2020Panoptic}
Bowen Cheng, Maxwell~D Collins, Yukun Zhu, Ting Liu, Thomas~S Huang, Hartwig
  Adam, and Liang-Chieh Chen.
\newblock Panoptic-{{DeepLab}}: {{A Simple}}, {{Strong}}, and {{Fast Baseline}}
  for {{Bottom-Up Panoptic Segmentation}}.
\newblock In {\em Proceedings of the {{IEEE}}/{{CVF Conference}} on {{Computer
  Vision}} and {{Pattern Recognition}} ({{CVPR}})}, pages 12475--12485, June
  2020.

\bibitem{Cheng2021Mask2Former}
Bowen Cheng, Ishan Misra, Alexander~G. Schwing, Alexander Kirillov, and Rohit
  Girdhar.
\newblock Masked-attention {{Mask Transformer}} for {{Universal Image
  Segmentation}}.
\newblock In {\em 2022 {{IEEE}}/{{CVF Conference}} on {{Computer Vision}} and
  {{Pattern Recognition}} ({{CVPR}})}, pages 1280--1289, 2022.

\bibitem{Cheng2021Are}
Yuwei Cheng, Mengxin Jiang, Jiannan Zhu, and Yimin Liu.
\newblock Are {{We Ready}} for {{Unmanned Surface Vehicles}} in {{Inland
  Waterways}}? {{The USVInland Multisensor Dataset}} and {{Benchmark}}.
\newblock {\em IEEE Robotics and Automation Letters}, 6(2):3964--3970, 2021.

\bibitem{Cheng2021FloW}
Yuwei Cheng, Jiannan Zhu, Mengxin Jiang, Jie Fu, Changsong Pang, Peidong Wang,
  Kris Sankaran, Olawale Onabola, Yimin Liu, Dianbo Liu, and Yoshua Bengio.
\newblock {{FloW}}: {{A Dataset}} and {{Benchmark}} for {{Floating Waste
  Detection}} in {{Inland Waters}}.
\newblock In {\em Proceedings of the {{IEEE}}/{{CVF International Conference}}
  on {{Computer Vision}} ({{ICCV}})}, pages 10953--10962, 2021.

\bibitem{mmseg2020}
{\relax Mms}egmentation Contributors.
\newblock {{MMSegmentation}}: {{OpenMMLab}} semantic segmentation toolbox and
  benchmark, 2020.

\bibitem{Cordts2016Cityscapes}
Marius Cordts, Mohamed Omran, Sebastian Ramos, Timo Rehfeld, Markus Enzweiler,
  Rodrigo Benenson, Uwe Franke, Stefan Roth, and Bernt Schiele.
\newblock The {{Cityscapes Dataset}} for {{Semantic Urban Scene
  Understanding}}.
\newblock In {\em Proceedings of the {{IEEE Conference}} on {{Computer Vision}}
  and {{Pattern Recognition}} ({{CVPR}})}, June 2016.

\bibitem{DeFilippo2021RoboWhaler}
Michael DeFilippo, Michael Sacarny, and Paul Robinette.
\newblock {{RoboWhaler}}: {{A Robotic Vessel}} for {{Marine Autonomy}} and
  {{Dataset Collection}}.
\newblock In {\em {{OCEANS}} 2021: {{San Diego}} \textendash{} {{Porto}}},
  pages 1--7, Sept. 2021.

\bibitem{Deng2020RetinaFace}
Jiankang Deng, Jia Guo, Evangelos Ververas, Irene Kotsia, and Stefanos
  Zafeiriou.
\newblock {{RetinaFace}}: {{Single-Shot Multi-Level Face Localisation}} in the
  {{Wild}}.
\newblock In {\em Proceedings of the {{IEEE}}/{{CVF Conference}} on {{Computer
  Vision}} and {{Pattern Recognition}}}, pages 5203--5212, 2020.

\bibitem{Fan2021Rethinking}
Mingyuan Fan, Shenqi Lai, Junshi Huang, Xiaoming Wei, Zhenhua Chai, Junfeng
  Luo, and Xiaolin Wei.
\newblock Rethinking {{BiSeNet For Real-time Semantic Segmentation}}.
\newblock In {\em CVPR 2021}, pages 9716--9725, Apr. 2021.

\bibitem{Gebru2021Datasheets}
Timnit Gebru, Jamie Morgenstern, Briana Vecchione, Jennifer~Wortman Vaughan,
  Hanna Wallach, Hal~Daum{\'e} III, and Kate Crawford.
\newblock Datasheets for datasets.
\newblock {\em Communications of the ACM}, 64(12):86--92, Nov. 2021.

\bibitem{Geiger2012Are}
Andreas Geiger, Philip Lenz, and Raquel Urtasun.
\newblock Are we ready for autonomous driving? {{The KITTI}} vision benchmark
  suite.
\newblock In {\em 2012 {{IEEE Conference}} on {{Computer Vision}} and {{Pattern
  Recognition}}}, pages 3354--3361, June 2012.

\bibitem{Kiefer20231st}
Benjamin Kiefer, Matej Kristan, Janez Per{\v s}, Lojze {\v Z}ust, Fabio Poiesi,
  Fabio Andrade, Alexandre Bernardino, Matthew Dawkins, Jenni Raitoharju,
  Yitong Quan, Adem Atmaca, Timon H{\"o}fer, Qiming Zhang, Yufei Xu, Jing
  Zhang, Dacheng Tao, Lars Sommer, Raphael Spraul, Hangyue Zhao, Hongpu Zhang,
  Yanyun Zhao, Jan~Lukas Augustin, Eui-ik Jeon, Impyeong Lee, Luca Zedda,
  Andrea Loddo, Cecilia Di~Ruberto, Sagar Verma, Siddharth Gupta, Shishir
  Muralidhara, Niharika Hegde, Daitao Xing, Nikolaos Evangeliou, Anthony Tzes,
  Vojt{\v e}ch Bartl, Jakub {\v S}pa{\v n}hel, Adam Herout, Neelanjan Bhowmik,
  Toby~P. Breckon, Shivanand Kundargi, Tejas Anvekar, Ramesh~Ashok Tabib, Uma
  Mudenagudi, Arpita Vats, Yang Song, Delong Liu, Yonglin Li, Shuman Li,
  Chenhao Tan, Long Lan, Vladimir Somers, Christophe De~Vleeschouwer, Alexandre
  Alahi, Hsiang-Wei Huang, Cheng-Yen Yang, Jenq-Neng Hwang, Pyong-Kun Kim,
  Kwangju Kim, Kyoungoh Lee, Shuai Jiang, Haiwen Li, Zheng Ziqiang, Tuan-Anh
  Vu, Hai {Nguyen-Truong}, Sai-Kit Yeung, Zhuang Jia, Sophia Yang, Chih-Chung
  Hsu, Xiu-Yu Hou, Yu-An Jhang, Simon Yang, and Mau-Tsuen Yang.
\newblock 1st {{Workshop}} on {{Maritime Computer Vision}} ({{MaCVi}}) 2023:
  {{Challenge Results}}.
\newblock In {\em Proceedings of the {{IEEE}}/{{CVF Winter Conference}} on
  {{Applications}} of {{Computer Vision}}}, pages 265--302, 2023.

\bibitem{Kirillov2019Panoptica}
Alexander Kirillov, Ross Girshick, Kaiming He, and Piotr Doll{\'a}r.
\newblock Panoptic {{Feature Pyramid Networks}}.
\newblock In {\em CVPR 2019}, pages 6399--6408, Apr. 2019.

\bibitem{Kirillov2019Panoptic}
Alexander Kirillov, Kaiming He, Ross Girshick, Carsten Rother, and Piotr
  Dollar.
\newblock Panoptic segmentation.
\newblock In {\em Proceedings of the {{IEEE Computer Society Conference}} on
  {{Computer Vision}} and {{Pattern Recognition}}}, volume 2019-June, pages
  9396--9405. {IEEE Computer Society}, June 2019.

\bibitem{Kirillov2020PointRend}
Alexander Kirillov, Yuxin Wu, Kaiming He, and Ross Girshick.
\newblock {{PointRend}}: {{Image Segmentation As Rendering}}.
\newblock In {\em 2020 {{IEEE}}/{{CVF Conference}} on {{Computer Vision}} and
  {{Pattern Recognition}} ({{CVPR}})}, pages 9796--9805, June 2020.

\bibitem{Kristan2016Fast}
Matej Kristan, Vildana~Suli{\'c} Kenk, Stanislav Kova{\v c}i{\v c}, and Janez
  Per{\v s}.
\newblock Fast {{Image-Based Obstacle Detection}} from {{Unmanned Surface
  Vehicles}}.
\newblock {\em IEEE Transactions on Cybernetics}, 46(3), 2016.

\bibitem{Kristan2015Graphical}
Matej Kristan, Janez Per{\v s}, Vildana Suli{\v c}, and Stanislav Kova{\v
  c}i{\v c}.
\newblock A {{Graphical Model}} for {{Rapid Obstacle Image-Map Estimation}}
  from {{Unmanned Surface Vehicles}}.
\newblock In Daniel Cremers, Ian Reid, Hideo Saito, and Ming-Hsuan Yang,
  editors, {\em Computer {{Vision}} -- {{ACCV}} 2014}, Lecture {{Notes}} in
  {{Computer Science}}, pages 391--406, {Cham}, 2015. {Springer International
  Publishing}.

\bibitem{Lambert2022ROSEBUD}
Reeve Lambert, Jalil {Chavez-Galaviz}, Jianwen Li, and Nina Mahmoudian.
\newblock {{ROSEBUD}}: {{A Deep Fluvial Segmentation Dataset}} for {{Monocular
  Vision-Based River Navigation}} and {{Obstacle Avoidance}}.
\newblock {\em Sensors}, 22(13):4681, June 2022.

\bibitem{Lin2014COCO}
Tsung~Yi Lin, Michael Maire, Serge Belongie, James Hays, Pietro Perona, Deva
  Ramanan, Piotr Doll{\'a}r, and C.~Lawrence Zitnick.
\newblock Microsoft {{COCO}}: {{Common}} objects in context.
\newblock In {\em LNCS}, volume 8693 LNCS, pages 740--755. {Springer Verlag},
  May 2014.

\bibitem{Lis2019Detecting}
Krzysztof Lis, Krishna~Kanth Nakka, Pascal Fua, and Mathieu Salzmann.
\newblock Detecting the {{Unexpected}} via {{Image Resynthesis}}.
\newblock In {\em ICCV 2019}, pages 2152--2161, {Seoul, Korea (South)}, Oct.
  2019. {IEEE}.

\bibitem{Long2015Fully}
Jonathan Long, Evan Shelhamer, and Trevor Darrell.
\newblock Fully convolutional networks for semantic segmentation.
\newblock In {\em Proceedings of the {{IEEE Computer Society Conference}} on
  {{Computer Vision}} and {{Pattern Recognition}}}, volume 07-12-June, pages
  431--440, 2015.

\bibitem{Ma2020Convolutional}
Liyong Ma, Wei Xie, and Haibin Huang.
\newblock Convolutional neural network based obstacle detection for unmanned
  surface vehicle.
\newblock {\em Mathematical Biosciences and Engineering}, 17(1), 2020.

\bibitem{Muhovic2020Obstacle}
Jon Muhovi{\v c}, Rok Mandeljc, Borja Bovcon, Matej Kristan, and Janez Per{\v
  s}.
\newblock Obstacle {{Tracking}} for {{Unmanned Surface Vessels Using}} 3-{{D
  Point Cloud}}.
\newblock {\em IEEE Journal of Oceanic Engineering}, 45(3), 2020.

\bibitem{Nirgudkar2022MassMIND}
Shailesh Nirgudkar, Michael DeFilippo, Michael Sacarny, Michael Benjamin, and
  Paul Robinette.
\newblock {{MassMIND}}: {{Massachusetts Maritime INfrared Dataset}}, Sept.
  2022.

\bibitem{Nirgudkar2021Visible}
Shailesh Nirgudkar and Paul Robinette.
\newblock Beyond {{Visible Light}}: {{Usage}} of {{Long Wave Infrared}} for
  {{Object Detection}} in {{Maritime Environment}}.
\newblock In {\em 2021 20th {{International Conference}} on {{Advanced
  Robotics}} ({{ICAR}})}, pages 1093--1100, Dec. 2021.

\bibitem{Prasad2019Object}
Dilip~K. Prasad, Chandrashekar~Krishna Prasath, Deepu Rajan, Lily Rachmawati,
  Eshan Rajabally, and Chai Quek.
\newblock Object {{Detection}} in a {{Maritime Environment}}: {{Performance
  Evaluation}} of {{Background Subtraction Methods}}.
\newblock {\em IEEE Transactions on Intelligent Transportation Systems},
  20(5):1787--1802, May 2019.

\bibitem{Prasad2017Video}
Dilip~K. Prasad, Deepu Rajan, Lily Rachmawati, Eshan Rajabally, and Chai Quek.
\newblock Video {{Processing From Electro-Optical Sensors}} for {{Object
  Detection}} and {{Tracking}} in a {{Maritime Environment}}: {{A Survey}}.
\newblock {\em IEEE Transactions on Intelligent Transportation Systems}, 18(8),
  2017.

\bibitem{Qiao2022Automated}
Dalei Qiao, Guangzhong Liu, Wei Li, Taizhi Lyu, and Juan Zhang.
\newblock Automated {{Full Scene Parsing}} for {{Marine ASVs Using Monocular
  Vision}}.
\newblock {\em Journal of Intelligent \& Robotic Systems}, 104(2):1--20, 2022.

\bibitem{Ren2017Faster}
Shaoqing Ren, Kaiming He, Ross Girshick, and Jian Sun.
\newblock Faster {{R-CNN}}: {{Towards Real-Time Object Detection}} with
  {{Region Proposal Networks}}.
\newblock {\em IEEE Transactions on Pattern Analysis and Machine Intelligence},
  39(6):1137--1149, June 2017.

\bibitem{Robinette2019Sensor}
Paul Robinette, Michael Sacarny, Michael Defilippo, Michael Novitzky, and
  Michael~R. Benjamin.
\newblock Sensor {{Evaluation}} for {{Autonomous Surface Vehicles}} in {{Inland
  Waterways}}.
\newblock In {\em {{OCEANS}} 2019 - {{Marseille}}, {{OCEANS Marseille}} 2019},
  volume 2019-June. {Institute of Electrical and Electronics Engineers Inc.},
  June 2019.

\bibitem{Ronneberger2015UNet}
Olaf Ronneberger, Philipp Fischer, and Thomas Brox.
\newblock U-net: {{Convolutional}} networks for biomedical image segmentation.
\newblock In {\em International {{Conference}} on {{Medical}} Image Computing
  and Computer-Assisted Intervention}, volume 9351, pages 234--241, 2015.

\bibitem{Steccanella2020}
L. Steccanella, D.~D. Bloisi, A. Castellini, and A. Farinelli.
\newblock Waterline and obstacle detection in images from low-cost autonomous
  boats for environmental monitoring.
\newblock {\em Robotics and Autonomous Systems}, 124, 2020.

\bibitem{Strudel2021Segmenter}
Robin Strudel, Ricardo Garcia, Ivan Laptev, and Cordelia Schmid.
\newblock Segmenter: {{Transformer}} for {{Semantic Segmentation}}.
\newblock In {\em Proceedings of the {{IEEE}}/{{CVF International Conference}}
  on {{Computer Vision}} ({{ICCV}})}, pages 7262--7272, May 2021.

\bibitem{Taipalmaa2019HighResolution}
Jussi Taipalmaa, Nikolaos Passalis, Honglei Zhang, Moncef Gabbouj, and Jenni
  Raitoharju.
\newblock High-{{Resolution Water Segmentation}} for {{Autonomous Unmanned
  Surface Vehicles}}: A {{Novel Dataset}} and {{Evaluation}}.
\newblock In {\em 2019 {{IEEE}} 29th {{International Workshop}} on {{Machine
  Learning}} for {{Signal Processing}} ({{MLSP}})}, pages 1--6, {Pittsburgh,
  PA, USA}, Oct. 2019. {IEEE}.

\bibitem{Tan2019EfficientDet}
Mingxing Tan, Ruoming Pang, and Quoc~V. Le.
\newblock {{EfficientDet}}: {{Scalable}} and {{Efficient Object Detection}}.
\newblock {\em Proceedings of the IEEE Computer Society Conference on Computer
  Vision and Pattern Recognition}, pages 10778--10787, Nov. 2019.

\bibitem{Tian2019FCOS}
Zhi Tian, Chunhua Shen, Hao Chen, and Tong He.
\newblock {{FCOS}}: {{Fully}} convolutional one-stage object detection.
\newblock In {\em Proceedings of the {{IEEE International Conference}} on
  {{Computer Vision}}}, volume 2019-Octob, 2019.

\bibitem{Wang2021Temporal}
Hao Wang, Weining Wang, and Jing Liu.
\newblock Temporal {{Memory Attention}} for {{Video Semantic Segmentation}}.
\newblock In {\em 2021 {{IEEE International Conference}} on {{Image
  Processing}} ({{ICIP}})}, pages 2254--2258. {IEEE}, Sept. 2021.

\bibitem{Wang2013Stereovision}
Han Wang and Zhuo Wei.
\newblock Stereovision based obstacle detection system for unmanned surface
  vehicle.
\newblock In {\em 2013 {{IEEE International Conference}} on {{Robotics}} and
  {{Biomimetics}}, {{ROBIO}} 2013}, 2013.

\bibitem{Wang2020MaXDeepLab}
Huiyu Wang, Yukun Zhu, Hartwig Adam, Alan Yuille, and Liang-Chieh Chen.
\newblock {{MaX-DeepLab}}: {{End-to-End Panoptic Segmentation}} with {{Mask
  Transformers}}.
\newblock In {\em Proceedings of the {{IEEE}}/{{CVF Conference}} on {{Computer
  Vision}} and {{Pattern Recognition}} ({{CVPR}})}, pages 5463--5474, Dec.
  2020.

\bibitem{Xie2021SegFormer}
Enze Xie, Wenhai Wang, Zhiding Yu, Anima Anandkumar, Jose~M. Alvarez, and Ping
  Luo.
\newblock {{SegFormer}}: {{Simple}} and {{Efficient Design}} for {{Semantic
  Segmentation}} with {{Transformers}}.
\newblock In {\em Advances in {{Neural Information Processing Systems}}},
  volume~34, pages 12077--12090, May 2021.

\bibitem{Yao2021Shoreline}
L Yao, D Kanoulas, Z Ji, and Y Liu.
\newblock {{ShorelineNet}}: {{An Efficient Deep Learning Approach}} for
  {{Shoreline Semantic Segmentation}} for {{Unmanned Surface Vehicles}}.
\newblock In {\em Proceedings of the 2021 {{IEEE}}/{{RSJ International
  Conference}} on {{Intelligent Robots}} and {{Systems}} ({{IROS}})}, 2021.

\bibitem{Yu2021BiSeNet}
Changqian Yu, Changxin Gao, Jingbo Wang, Gang Yu, Chunhua Shen, and Nong Sang.
\newblock {{BiSeNet V2}}: {{Bilateral Network}} with {{Guided Aggregation}} for
  {{Real-Time Semantic Segmentation}}.
\newblock {\em International Journal of Computer Vision}, 129(11):3051--3068,
  Nov. 2021.

\bibitem{Yu2018Bisenet}
Changqian Yu, Jingbo Wang, Chao Peng, Changxin Gao, Gang Yu, and Nong Sang.
\newblock {{BiSeNet}}: {{Bilateral}} segmentation network for real-time
  semantic segmentation.
\newblock In {\em Computer {{Vision}} - {{ECCV}} 2018}, pages 334--349, Aug.
  2018.

\bibitem{Yuan2021CSANet}
Yichen Yuan, Lijun Wang, and Yifan Wang.
\newblock {{CSANet}} for {{Video Semantic Segmentation With Inter-Frame Mutual
  Learning}}.
\newblock {\em IEEE Signal Processing Letters}, 28:1675--1679, 2021.

\bibitem{Zendel2022Unifying}
Oliver Zendel, Matthias Sch{\"o}rghuber, Bernhard Rainer, Markus Murschitz, and
  Csaba Beleznai.
\newblock Unifying {{Panoptic Segmentation}} for {{Autonomous Driving}}.
\newblock In {\em 2022 {{IEEE}}/{{CVF Conference}} on {{Computer Vision}} and
  {{Pattern Recognition}} ({{CVPR}})}, pages 21319--21328, June 2022.

\bibitem{Zhang2021KNet}
Wenwei Zhang, Jiangmiao Pang, Kai Chen, and Chen~Change Loy.
\newblock K-{{Net}}: {{Towards Unified Image Segmentation}}.
\newblock In {\em Advances in {{Neural Information Processing Systems}}}, Oct.
  2021.

\bibitem{Zust2022Temporal}
Lojze {\v Z}ust and Matej Kristan.
\newblock Temporal {{Context}} for {{Robust Maritime Obstacle Detection}}.
\newblock In {\em 2022 {{IEEE}}/{{RJS International Conference}} on
  {{Intelligent Robots}} and {{Systems}} ({{IROS}})}, 2022.

\end{thebibliography}
}

\newpage
\appendix

\section{Obstacle heatmaps}
\setcounter{figure}{0}

\begin{figure}
  \centering
  \includegraphics[width=\linewidth]{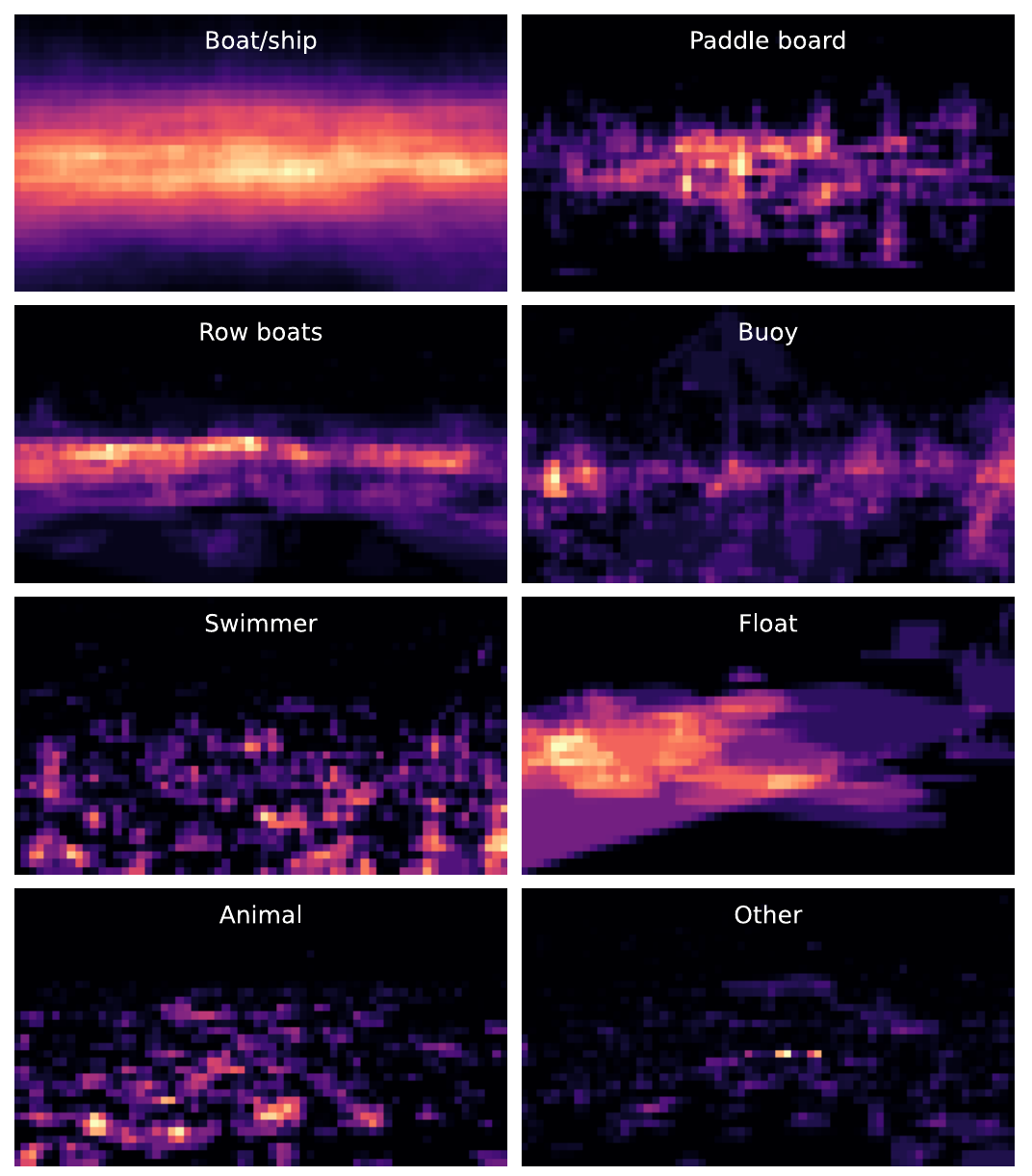}
  \caption{Heatmaps of obstacle categories. Dark and bright colors denote areas with low and high frequency of obstacles, respectively.}
    \label{fig:obst-heatmap}
\end{figure}

To visualize the spatial distribution of obstacles of different categories, we compute heatmaps of obstacle positions. We divide the image into $64 \times 36$ spatial bins. In each bin we count the number of obstacle segments which intersect with the bin. The heatmaps in Figure~\ref{fig:obst-heatmap} are normalized with respect to the most populated bin in each category.

Most categories are evenly distributed across the image. Various vessel categories (\eg boat/ship, row boats) vertically most commonly appear near the center of the image, which coincides with the usual position of the horizon. Some smaller categories such as swimmer, animal and buoy contain many instances closer to the camera as well.

\section{Additional semantic segmentation results}
\setcounter{figure}{0}

\begin{figure}
  \centering
  \includegraphics[width=\linewidth]{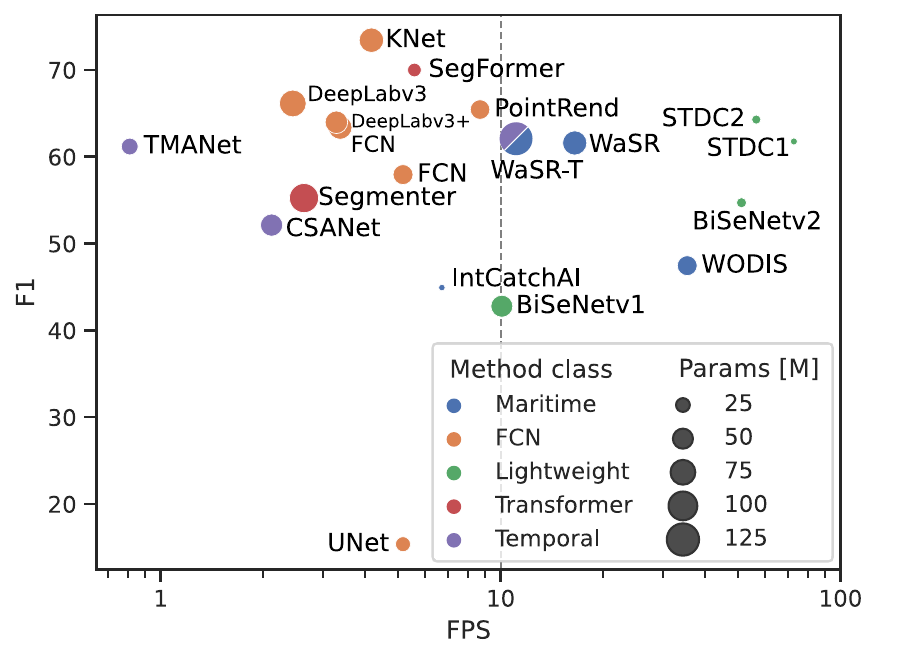}
  \caption{The performance of semantic segmentation method (F1) with respect to their efficiency, measured in FPS. Dashed line denotes the real-time boundary of 10 FPS.}
    \label{fig:seg-eff}
\end{figure} 

\subsection{Method detection efficiency}

To better understand the trade-offs between detection performance and speed we plot the obstacle detection F1 score of methods with respect to their inference time measured in FPS in Figure~\ref{fig:seg-eff}. Most methods do not reach the real-time inference speed requirement of 10 FPS including top-performing KNet~\cite{Zhang2021KNet} and SegFormer~\cite{Xie2021SegFormer}. WaSR-T~\cite{Zust2022Temporal} and WaSR~\cite{Bovcon2021WaSR} both perform on the limit of this requirement and achieve an F1 score of over 60. Among the real-time methods, the STDC~\cite{Fan2021Rethinking} family offers the best trade-off between speed and performance by a large margin achieving best results in both.

\subsection{Additional qualitative results}

\begin{figure*}
  \centering
  \includegraphics[width=\linewidth]{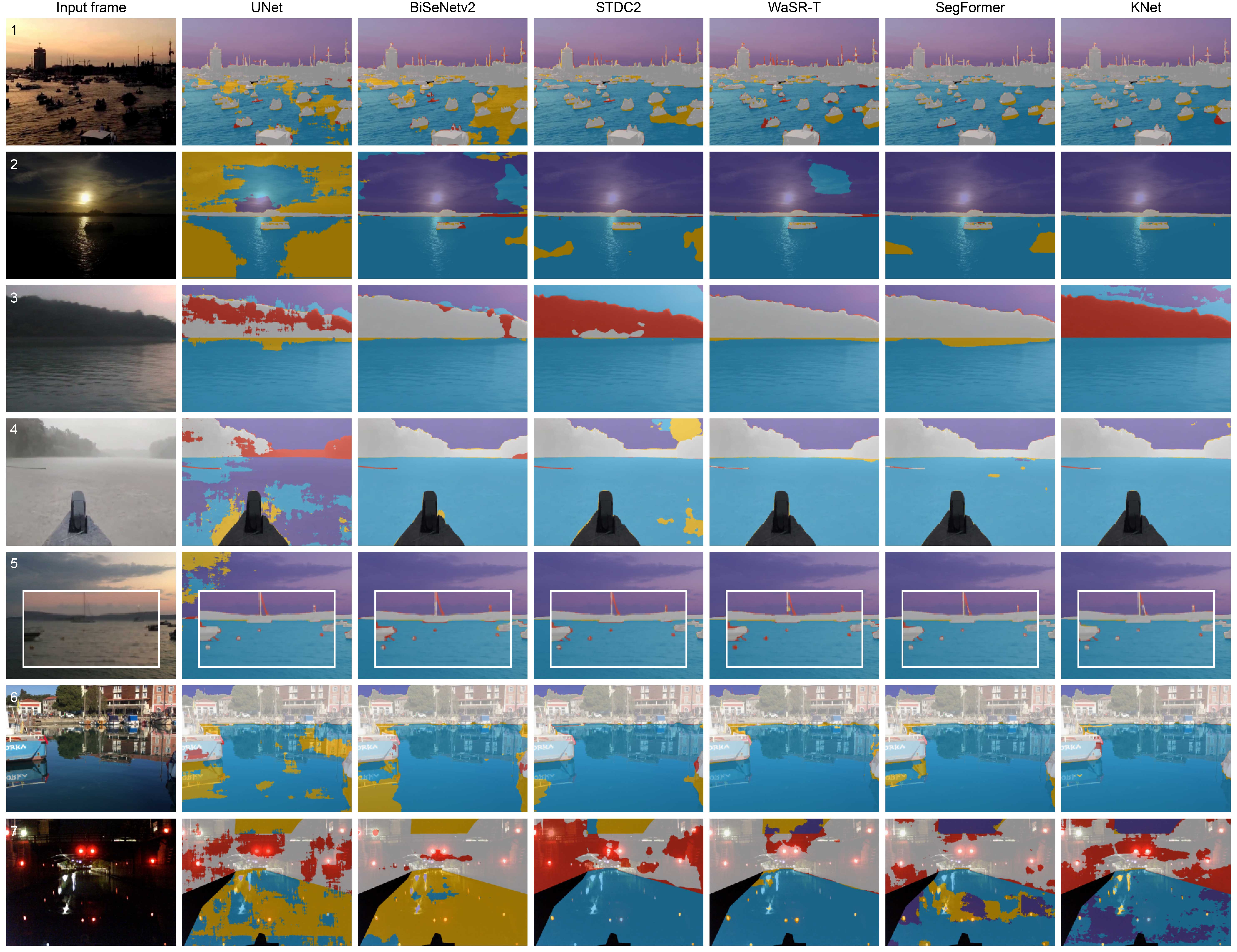}
  \caption{Additional qualitative semantic segmentation results. Sky and water classes are shown in purple and blue, respectively. TP, FN and FP obstacle predictions are shown in white, red and yellow, respectively, while black indicates ignore region. White rectangles show zoomed-in parts of the image.}
    \label{fig:semantic-qualitative-add}
\end{figure*}

Figure~\ref{fig:semantic-qualitative-add} showcases additional qualitative results for semantic segmentation methods, including low-visibility and night scenes (rows 1, 2 and 7), foggy and rainy scenes (rows 3 and 4), small obstacles (row 5) and reflections (row 6). Methods like UNet~\cite{Ronneberger2015UNet}, BiSeNetv2~\cite{Yu2021BiSeNet} and SegFormer~\cite{Xie2021SegFormer} are prone to obstacle hallucinations in high-ambiguity scenes (rows 2, 6 and 7). WaSR-T~\cite{Zust2022Temporal} and KNet~\cite{Zhang2021KNet} are the most robust to these ambiguities. Small obstacles (row 5) are only picked up by SegFormer~\cite{Xie2021SegFormer} and KNet~\cite{Zhang2021KNet}. Interestingly, even the best performing methods (\eg KNet~\cite{Zhang2021KNet}) sometimes fail in seemingly simple situations such as the scene in row 3, where the slight ambient fog leads to complete misclassification of the large landmass as water in several methods.

\section{Additional panoptic segmentation results}
\setcounter{figure}{0}

\subsection{Performance by obstacle size}

\begin{figure}
  \centering
  \includegraphics[width=\linewidth]{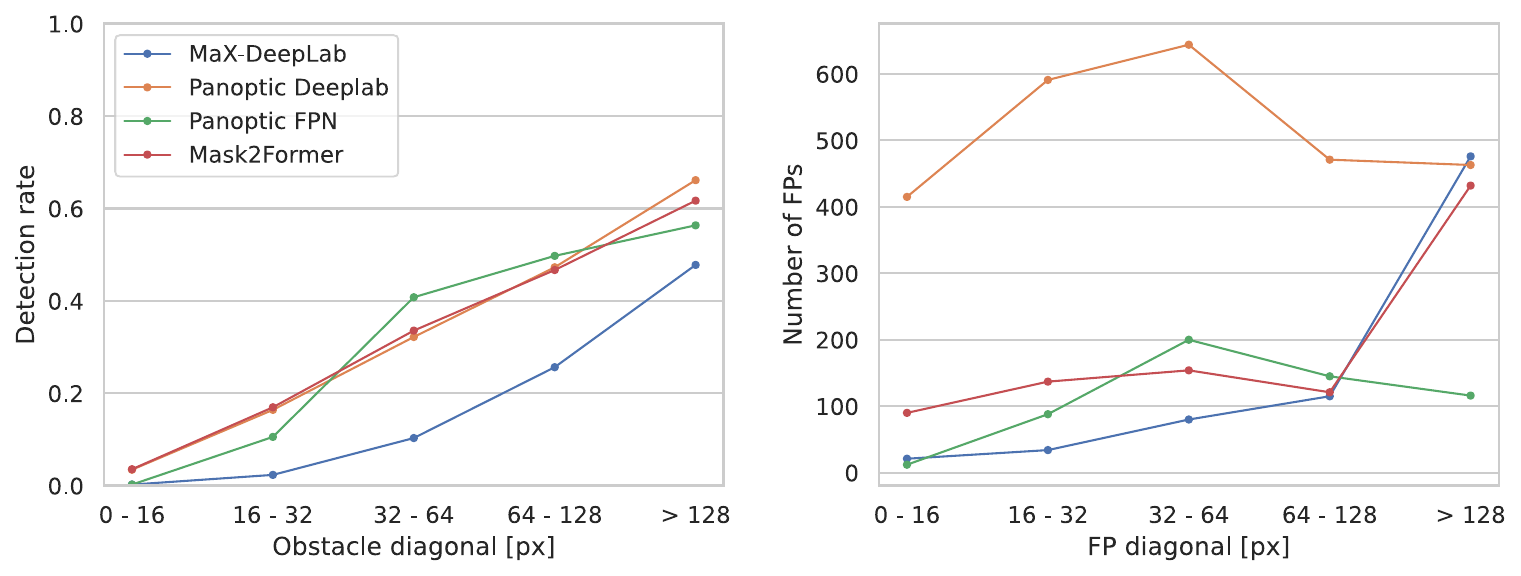}
  \caption{Performance (PQ) of panoptic methods by obstacle size.}
    \label{fig:pan-size}
\end{figure} 

Figure~\ref{fig:pan-size} shows the performance of panoptic methods (PQ) with respect to the obstacle size. Similarly to semantic segmentation methods, the best performance is observed on large obstacles. However, on very small obstacles, the performance drops to almost zero. We believe there is a large potential for improving panoptic methods in this regard. Note also, that the PQ metric is much more sensitive to minor mask shifts on small obstacles compared to large ones, which also impacts these results. In contrast to semantic segmentation methods, large false-positive detections are more common than small ones, with the exception of Panoptic DeepLab, which also produces a large number of smaller false-positive detections.

\subsection{Source of detection errors}

\begin{figure}
  \centering
  \includegraphics[width=\linewidth]{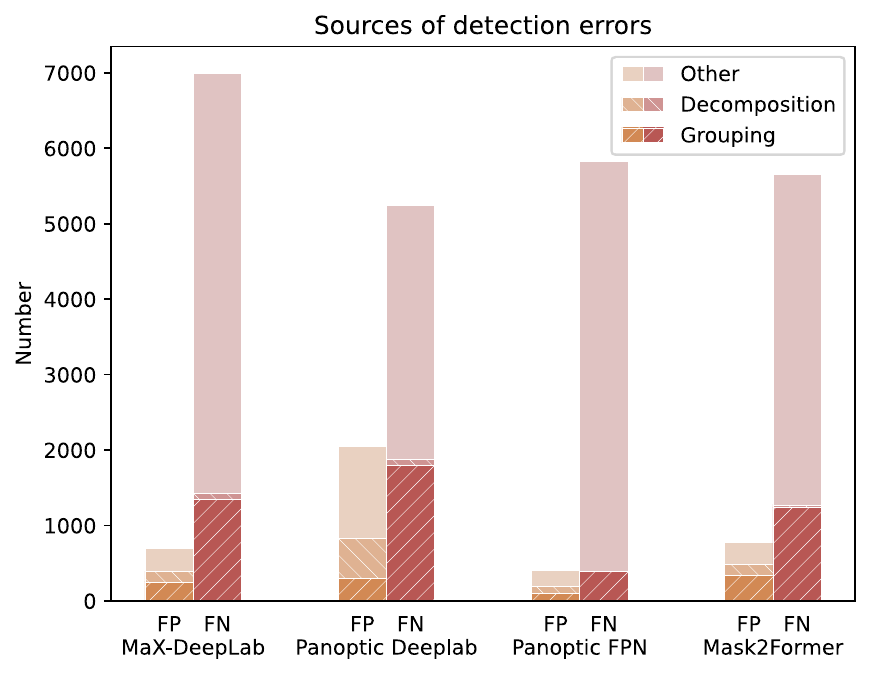}
  \caption{Common sources of FP and FN errors for panoptic detection methods.}
    \label{fig:pan-errors}
\end{figure} 

To further explore the problem of object grouping and object decomposition errors discussed in Section~5.2, we investigate the frequency of these problems across the different methods. Specifically, we inspect the source of false-positive and false-negative detections in the obstacle-class-agnostic case. A FP segment is counted as result of a decomposition error, if it is largely (more than 70\% of its area) contained within a single ground-truth obstacle segment. A FP segment is similarly counted as a result of a grouping error if it covers more than one ground-truth obstacle segments. A FN ground-truth segment is counted as a result of decomposition, if the combined coverage of all predicted obstacle segments exceeds the threshold of 50\%, and as a result of grouping if there exists any single predicted segment, that that covers more than 50\% of the FN segment.

The proportion of the two sources of errors for each method are reported in Figure~\ref{fig:pan-errors}. We observe that a sizeable amount of FP and FN detections are the result of these errors. Object grouping is an especially common source of false-negative detections. Addressing the issue of object grouping would thus lead to substantial performance improvements.

\subsection{Obstacle confusion matrices}

\begin{figure}
  \centering
  \includegraphics[width=\linewidth]{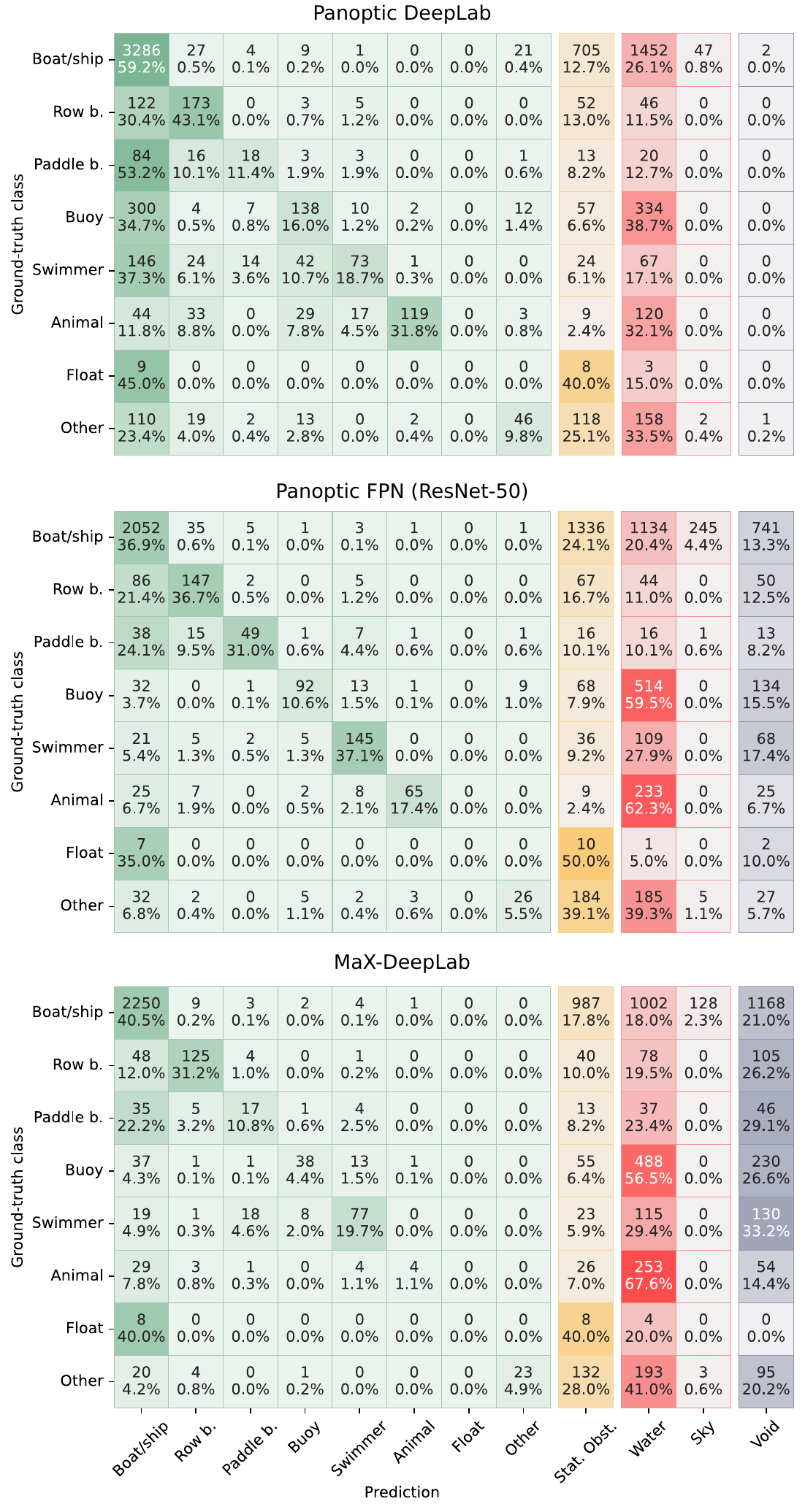}
  \caption{Ground-truth dynamic obstacle confusion matrices for panoptic methods.}
    \label{fig:comb-confm}
\end{figure} 

Similarly to the Figure 8 of the main paper, we plot the confusion matrices for the remaining panoptic methods (Panoptic DeepLab~\cite{Cheng2020Panoptic}, Panoptic FPN~\cite{Kirillov2019Panoptica} and MaX-DeepLab~\cite{Wang2020MaXDeepLab}) in Figure~\ref{fig:comb-confm}. Compared to Mask2Former~\cite{Cheng2021Mask2Former}, Panoptic DeepLab~\cite{Cheng2020Panoptic} does not rely on void predictions and correctly identifies more obstacle instances. However, the confusion between individual obstacle types is much larger. For example, Panoptic DeepLab~\cite{Cheng2020Panoptic} tends to classify most obstacles to the majority \textit{boat/ship} category. On the other hand, Panoptic FPN~\cite{Kirillov2019Panoptica} and MaX-DeepLab~\cite{Wang2020MaXDeepLab} show concerning level of misclassifications of obstacles as water, which is potentially hazardous from the boat navigation perspective. This problem is especially prevalent on smaller obstacle categories such as buoys and animals.

\subsection{Additional qualitative results}

\begin{figure*}[p]
  \centering
  \includegraphics[width=\linewidth]{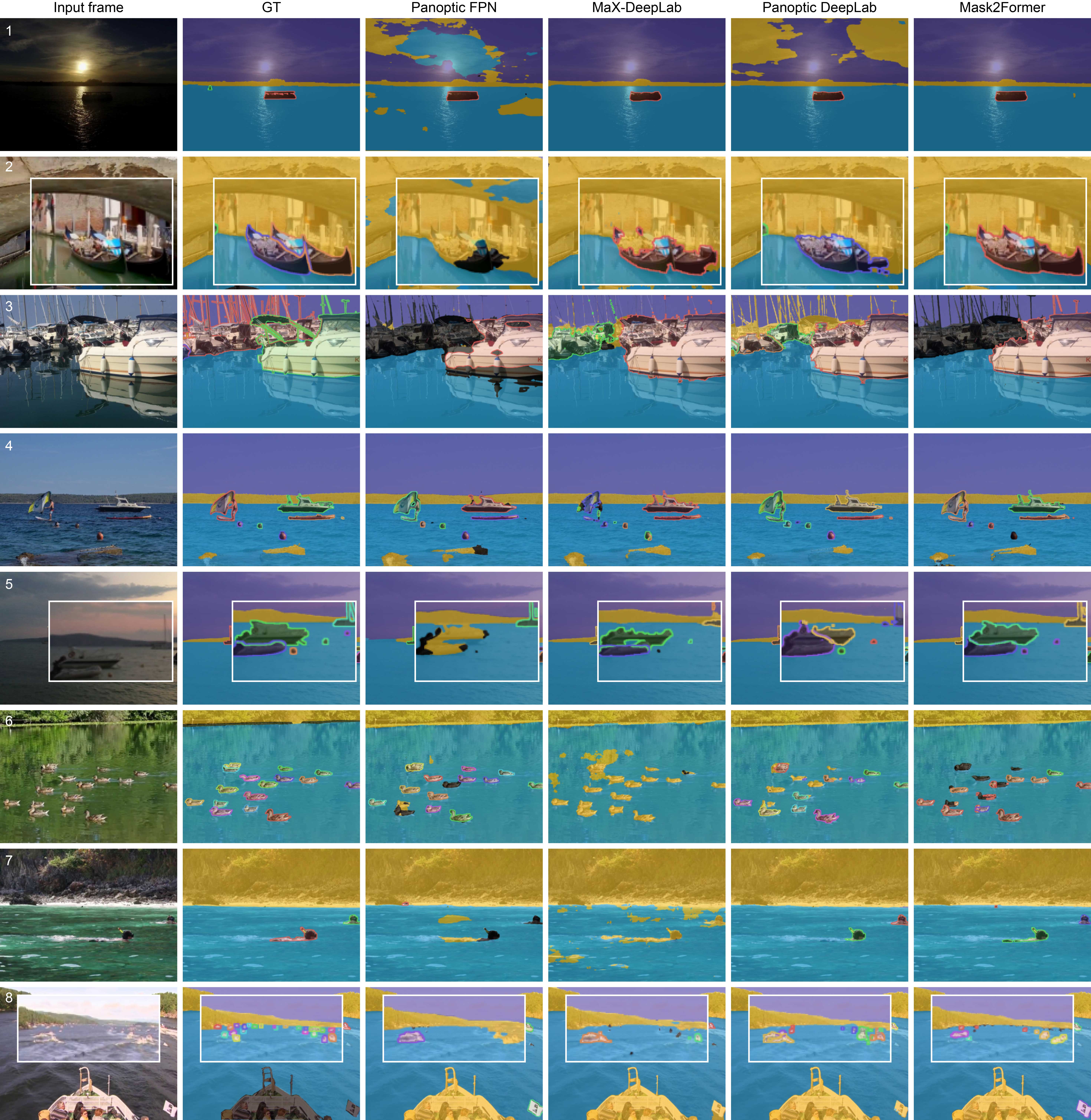}
  \caption{Additional qualitative panoptic results. Individual instance detections are outlined with different colors. Void predictions are colored black. White rectangles show zoomed-in parts of the image.}
    \label{fig:panoptic-qualitative-add}
\end{figure*}

Figure~\ref{fig:panoptic-qualitative-add} showcases additional qualitative results for panoptic segmentation methods. Small or far-away objects (rows 5 and 8) are often missed (Panoptic FPN~\cite{Kirillov2019Panoptica} and Panoptic DeepLab~\cite{Cheng2020Panoptic}) or labeled as void (MaX-DeepLab~\cite{Wang2020MaXDeepLab} and Mask2Former~\cite{Cheng2021Mask2Former}). Similar objects that are close together (rows 2, 5 and 6) are commonly grouped as a single object. Additionally, Mask2Former~\cite{Cheng2021Mask2Former} sometimes groups even far-away objects (\eg buoys in row 5 and ducks in row 6).

\clearpage
\newpage

\section{Datasheet for LaRS}

\noindent
\textbf{
This document is based on \textit{Datasheets for Datasets} by Gebru \textit{et
al.}~\cite{Gebru2021Datasheets}. Please see the most updated version
\underline{\textcolor{blue}{\href{http://arxiv.org/abs/1803.09010}{here}}}.
}

{\small

\begin{mdframed}[linecolor=datasheet]
\textbf{\textcolor{datasheet}{
    MOTIVATION
}}
\end{mdframed}

    \textcolor{datasheet}{\textbf{
    For what purpose was the dataset created?
    }
    Was there a specific task in mind? Was there
    a specific gap that needed to be filled? Please provide a description.
    } \\
    LaRS was created as a benchmark for panoptic maritime obstacle detection, to facilitate the development and evaluation of new panoptic (and semantic) segmentation methods for robust obstacle detection under a wide range of conditions and situations. \\
    
    \textcolor{datasheet}{\textbf{
    Who created this dataset (e.g., which team, research group) and on behalf
    of which entity (e.g., company, institution, organization)?
    }
    } \\
    The detaset was created by the ViCoS lab at the University of Ljubljana, Slovenia. \\
    
    \textcolor{datasheet}{\textbf{
    What support was needed to make this dataset?
    }
    (e.g.who funded the creation of the dataset? If there is an associated
    grant, provide the name of the grantor and the grant name and number, or if
    it was supported by a company or government agency, give those details.)
    } \\
    The creation of the dataset was funded by the Slovenian Research Agency program P2-0214 and project J2-2506.\\
    
    \textcolor{datasheet}{\textbf{
    Any other comments?
    }} \\
    No. \\

\begin{mdframed}[linecolor=datasheet]
\textbf{\textcolor{datasheet}{
    COMPOSITION
}}
\end{mdframed}
    \textcolor{datasheet}{\textbf{
    What do the instances that comprise the dataset represent (e.g., documents,
    photos, people, countries)?
    }
    Are there multiple types of instances (e.g., movies, users, and ratings;
    people and interactions between them; nodes and edges)? Please provide a
    description.
    } \\
    Instances in the dataset are snippets (i.e. scenes) of 10 sequential video frames (photos) depicting maritime scenarios captured from the perspective of a USV. \\
    
    \textcolor{datasheet}{\textbf{
    How many instances are there in total (of each type, if appropriate)?
    }
    } \\
    The dataset contains four thousand instances.\\
    
    \textcolor{datasheet}{\textbf{
    Does the dataset contain all possible instances or is it a sample (not
    necessarily random) of instances from a larger set?
    }
    If the dataset is a sample, then what is the larger set? Is the sample
    representative of the larger set (e.g., geographic coverage)? If so, please
    describe how this representativeness was validated/verified. If it is not
    representative of the larger set, please describe why not (e.g., to cover a
    more diverse range of instances, because instances were withheld or
    unavailable).
    } \\
    The instance were extracted from a larger set of videos. The videos were manually selected to feature diverse scenarios and geographic locations. At least one instance was extracted from each video to ensure visual diversity. Additional challenging instances were extracted through a visual inspection of predictions of a state-of-the-art (SotA) method. \\
    
    \textcolor{datasheet}{\textbf{
    What data does each instance consist of?
    }
    “Raw” data (e.g., unprocessed text or images) or features? In either case,
    please provide a description.
    } \\
    Each snippet contains 10 image frames. The image frames were processed to blur faces to protect the identities of individuals in the image. \\
    
    \textcolor{datasheet}{\textbf{
    Is there a label or target associated with each instance?
    }
    If so, please provide a description.
    } \\
    One \emph{"key"} video frame in the snippet is annotated with panoptic masks. This includes "water", "sky" and "static obstacle" stuff classes and 8 different dynamic obstacle categories (\textit{i.e.} things). The average image has $\sim$9 masks, totaling $\sim$36k masks. Each scene is also annotated with 19 different global attributes covering different environment types, reflection levels and other conditions. \\
    
    \textcolor{datasheet}{\textbf{
    Is any information missing from individual instances?
    }
    If so, please provide a description, explaining why this information is
    missing (e.g., because it was unavailable). This does not include
    intentionally removed information, but might include, e.g., redacted text.
    } \\
    The annotations of the test set will not be made publicly available to ensure fair comparison between methods. We host an evaluation server (\href{https://macvi.org}{macvi.org}) for submitting and evaluating the results of new methods. \\
    
    \textcolor{datasheet}{\textbf{
    Are relationships between individual instances made explicit (e.g., users’
    movie ratings, social network links)?
    }
    If so, please describe how these relationships are made explicit.
    } \\
    In some cases, several snippets have been extracted from a single video. We include the ID of the source sequence in the naming of the instance to make this relationship explicit. \\
    
    \textcolor{datasheet}{\textbf{
    Are there recommended data splits (e.g., training, development/validation,
    testing)?
    }
    If so, please provide a description of these splits, explaining the
    rationale behind them.
    } \\
    We provide a recommended data split into training (65 \%), validation (5 \%) and test (30 \%) set. Source sequences are mutually exclusive between sets. We insure equal distribution of resolution, reflection levels and scene types across sets. \\
    
    \textcolor{datasheet}{\textbf{
    Are there any errors, sources of noise, or redundancies in the dataset?
    }
    If so, please provide a description.
    } \\
    The annotations were created by human annotators and verified by us. Nonetheless, minor inconsistencies among different human annotators are possible. Annotation errors may be reported to \href{mailto:lars.dataset@gmail.com}{lars.dataset@gmail.com}. \\
    
    \textcolor{datasheet}{\textbf{
    Is the dataset self-contained, or does it link to or otherwise rely on
    external resources (e.g., websites, tweets, other datasets)?
    }
    If it links to or relies on external resources, a) are there guarantees
    that they will exist, and remain constant, over time; b) are there official
    archival versions of the complete dataset (i.e., including the external
    resources as they existed at the time the dataset was created); c) are
    there any restrictions (e.g., licenses, fees) associated with any of the
    external resources that might apply to a future user? Please provide
    descriptions of all external resources and any restrictions associated with
    them, as well as links or other access points, as appropriate.
    } \\
    The dataset is self-contained. \\
    
    \textcolor{datasheet}{\textbf{
    Does the dataset contain data that might be considered confidential (e.g.,
    data that is protected by legal privilege or by doctor-patient
    confidentiality, data that includes the content of individuals’ non-public
    communications)?
    }
    If so, please provide a description.
    } \\
    No. \\
    
    \textcolor{datasheet}{\textbf{
    Does the dataset contain data that, if viewed directly, might be offensive,
    insulting, threatening, or might otherwise cause anxiety?
    }
    If so, please describe why.
    } \\
    No. \\
    
    \textcolor{datasheet}{\textbf{
    Does the dataset relate to people?
    }
    If not, you may skip the remaining questions in this section.
    } \\
    No. \\
    
    \textcolor{datasheet}{\textbf{
    Does the dataset identify any subpopulations (e.g., by age, gender)?
    }
    If so, please describe how these subpopulations are identified and
    provide a description of their respective distributions within the dataset.
    } \\
    No. \\
    
    \textcolor{datasheet}{\textbf{
    Is it possible to identify individuals (i.e., one or more natural persons),
    either directly or indirectly (i.e., in combination with other data) from
    the dataset?
    }
    If so, please describe how.
    } \\
    No. We blur the faces of people appearing in the images to protect their identity. Issues with anonymization may be reported by email to \href{mailto:lars.dataset@gmail.com}{lars.dataset@gmail.com}. \\
    
    \textcolor{datasheet}{\textbf{
    Does the dataset contain data that might be considered sensitive in any way
    (e.g., data that reveals racial or ethnic origins, sexual orientations,
    religious beliefs, political opinions or union memberships, or locations;
    financial or health data; biometric or genetic data; forms of government
    identification, such as social security numbers; criminal history)?
    }
    If so, please provide a description.
    } \\
    No. \\
    
    \textcolor{datasheet}{\textbf{
    Any other comments?
    }} \\
    No. \\

\begin{mdframed}[linecolor=datasheet]
\textbf{\textcolor{datasheet}{
    COLLECTION
}}
\end{mdframed}

    \textcolor{datasheet}{\textbf{
    How was the data associated with each instance acquired?
    }
    Was the data directly observable (e.g., raw text, movie ratings),
    reported by subjects (e.g., survey responses), or indirectly
    inferred/derived from other data (e.g., part-of-speech tags, model-based
    guesses for age or language)? If data was reported by subjects or
    indirectly inferred/derived from other data, was the data
    validated/verified? If so, please describe how.
    } \\
    The instances in the dataset were collected from a combination of online sources (publicly available videos and datasets) and recordings from members of our lab. The corresponding panoptic masks were annotated by a professional labelling company and verified by us.\\
    
    \textcolor{datasheet}{\textbf{
    Over what timeframe was the data collected?
    }
    Does this timeframe match the creation timeframe of the data associated
    with the instances (e.g., recent crawl of old news articles)? If not,
    please describe the timeframe in which the data associated with the
    instances was created. Finally, list when the dataset was first published.
    } \\
    The instances in the dataset vary in their date of capture over a range of years up to 2023. The date of the first publication of the dataset is 1 August 2023. \\
    
    \textcolor{datasheet}{\textbf{
    What mechanisms or procedures were used to collect the data (e.g., hardware
    apparatus or sensor, manual human curation, software program, software
    API)?
    }
    How were these mechanisms or procedures validated?
    } \\
    The instances were captured with a wide range of different consumer-grade and industry-grade RGB cameras. \\
    
    \textcolor{datasheet}{\textbf{
    What was the resource cost of collecting the data?
    }
    (e.g. what were the required computational resources, and the associated
    financial costs, and energy consumption - estimate the carbon footprint.)
    } \\
    Since the sources of the dataset instances were pre-existing videos and videos captured during vacation time of our team members, no additional resource cost occurred during the collection process. \\
    
    \textcolor{datasheet}{\textbf{
    If the dataset is a sample from a larger set, what was the sampling
    strategy (e.g., deterministic, probabilistic with specific sampling
    probabilities)?
    }
    } \\
    Short snippets were extracted from longer video sequences. The selected snippets were determined manually based on the visual variety of the scene and difficulty, determined by the performance of existing obstacle detection methods. \\
    
    \textcolor{datasheet}{\textbf{
    Who was involved in the data collection process (e.g., students,
    crowdworkers, contractors) and how were they compensated (e.g., how much
    were crowdworkers paid)?
    }
    } \\
    The panoptic masks and category labels were annotated by a professional annotation service. The annotators were compensated with an hourly wage set by the vendor. \\
    
    \textcolor{datasheet}{\textbf{
    Were any ethical review processes conducted (e.g., by an institutional
    review board)?
    }
    If so, please provide a description of these review processes, including
    the outcomes, as well as a link or other access point to any supporting
    documentation.
    } \\
    We underwent an internal privacy review to evaluate and determine how to mitigate any potential risks with respect to the privacy of people appearing in the photos. Blurring faces protects the privacy of the people in the photos. \\
    
    \textcolor{datasheet}{\textbf{
    Does the dataset relate to people?
    }
    If not, you may skip the remainder of the questions in this section.
    } \\
    No. \\
    
    \textcolor{datasheet}{\textbf{
    Any other comments?
    }} \\
    No. \\

\begin{mdframed}[linecolor=datasheet]
\textbf{\textcolor{datasheet}{
    PREPROCESSING / CLEANING / LABELING
}}
\end{mdframed}

    \textcolor{datasheet}{\textbf{
    Was any preprocessing/cleaning/labeling of the data
    done(e.g.,discretization or bucketing, tokenization, part-of-speech
    tagging, SIFT feature extraction, removal of instances, processing of
    missing values)?
    }
    If so, please provide a description. If not, you may skip the remainder of
    the questions in this section.
    } \\
    We blur the faces to preserve the privacy of the individuals. No other preprocessing was done to the photos. \\

    \textcolor{datasheet}{\textbf{
    Was the “raw” data saved in addition to the preprocessed/cleaned/labeled
    data (e.g., to support unanticipated future uses)?
    }
    If so, please provide a link or other access point to the “raw” data.
    } \\
    No, because we preprocess the data to preserve the privacy of individuals, we do not release raw data. \\

    \textcolor{datasheet}{\textbf{
    Is the software used to preprocess/clean/label the instances available?
    }
    If so, please provide a link or other access point.
    } \\
    We used the RetinaFace model~\cite{Deng2020RetinaFace} (\url{https://github.com/serengil/retinaface}) to detect faces in the photos. \\

    \textcolor{datasheet}{\textbf{
    Any other comments?
    }} \\
    No. \\

\begin{mdframed}[linecolor=datasheet]
\textbf{\textcolor{datasheet}{
    USES
}}
\end{mdframed}

    \textcolor{datasheet}{\textbf{
    Has the dataset been used for any tasks already?
    }
    If so, please provide a description.
    } \\
    The dataset was used to train and evaluate 27 different semantic and panoptic segmentation methods. \\

    \textcolor{datasheet}{\textbf{
    Is there a repository that links to any or all papers or systems that use the dataset?
    }
    If so, please provide a link or other access point.
    } \\
    No. However, we require the users of the dataset to cite it in their papers, so its use is trackable via citations. \\

    \textcolor{datasheet}{\textbf{
    What (other) tasks could the dataset be used for?
    }
    } \\
    The dataset was intended for training and evaluation of semantic and panoptic segmentation methods. However, with minimal effort the dataset could also be used for other task such as instance segmentation and object detection.  \\

    \textcolor{datasheet}{\textbf{
    Is there anything about the composition of the dataset or the way it was
    collected and preprocessed/cleaned/labeled that might impact future uses?
    }
    For example, is there anything that a future user might need to know to
    avoid uses that could result in unfair treatment of individuals or groups
    (e.g., stereotyping, quality of service issues) or other undesirable harms
    (e.g., financial harms, legal risks) If so, please provide a description.
    Is there anything a future user could do to mitigate these undesirable
    harms?
    } \\
    We do not foresee any such impact of future uses. \\

    \textcolor{datasheet}{\textbf{
    Are there tasks for which the dataset should not be used?
    }
    If so, please provide a description.
    } \\
    Full terms of use for the dataset can be found at \url{https://lojzezust.github.io/lars-dataset}. \\

    \textcolor{datasheet}{\textbf{
    Any other comments?
    }} \\
    No. \\

\begin{mdframed}[linecolor=datasheet]
\textbf{\textcolor{datasheet}{
    DISTRIBUTION
}}
\end{mdframed}

    \textcolor{datasheet}{\textbf{
    Will the dataset be distributed to third parties outside of the entity
    (e.g., company, institution, organization) on behalf of which the dataset
    was created?
    }
    If so, please provide a description.
    } \\
    Yes, the dataset will be available to the research community. \\

    \textcolor{datasheet}{\textbf{
    How will the dataset will be distributed (e.g., tarball on website, API,
    GitHub)?
    }
    Does the dataset have a digital object identifier (DOI)?
    } \\
    The dataset is available at \url{https://lojzezust.github.io/lars-dataset} \\

    \textcolor{datasheet}{\textbf{
    When will the dataset be distributed?
    }
    } \\
    The dataset was released online on 1 August 2023 \\

    \textcolor{datasheet}{\textbf{
    Will the dataset be distributed under a copyright or other intellectual
    property (IP) license, and/or under applicable terms of use (ToU)?
    }
    If so, please describe this license and/or ToU, and provide a link or other
    access point to, or otherwise reproduce, any relevant licensing terms or
    ToU, as well as any fees associated with these restrictions.
    } \\
    The licence agreement and terms of use can be found at \url{https://lojzezust.github.io/lars-dataset} \\

    \textcolor{datasheet}{\textbf{
    Have any third parties imposed IP-based or other restrictions on the data
    associated with the instances?
    }
    If so, please describe these restrictions, and provide a link or other
    access point to, or otherwise reproduce, any relevant licensing terms, as
    well as any fees associated with these restrictions.
    } \\
    No. \\

    \textcolor{datasheet}{\textbf{
    Do any export controls or other regulatory restrictions apply to the
    dataset or to individual instances?
    }
    If so, please describe these restrictions, and provide a link or other
    access point to, or otherwise reproduce, any supporting documentation.
    } \\
    No. \\

    \textcolor{datasheet}{\textbf{
    Any other comments?
    }} \\
    No. \\

\begin{mdframed}[linecolor=datasheet]
\textbf{\textcolor{datasheet}{
    MAINTENANCE
}}
\end{mdframed}

    \textcolor{datasheet}{\textbf{
    Who is supporting/hosting/maintaining the dataset?
    }
    } \\
    The dataset will be hosted and maintained by the ViCoS lab, University of Ljubljana. \\

    \textcolor{datasheet}{\textbf{
    How can the owner/curator/manager of the dataset be contacted (e.g., email
    address)?
    }
    } \\
    Please email \href{mailto:lars.dataset@gmail.com}{lars.dataset@gmail.com}. \\

    \textcolor{datasheet}{\textbf{
    Is there an erratum?
    }
    If so, please provide a link or other access point.
    } \\
    No. \\

    \textcolor{datasheet}{\textbf{
    Will the dataset be updated (e.g., to correct labeling errors, add new
    instances, delete instances)?
    }
    If so, please describe how often, by whom, and how updates will be
    communicated to users (e.g., mailing list, GitHub)?
    } \\
    The dataset may be updated in case of discovered privacy concerns and major labeling errors. In this case, the version history and changes will be made clear on the dataset website (\url{https://lojzezust.github.io/lars-dataset}). \\

    \textcolor{datasheet}{\textbf{
    If the dataset relates to people, are there applicable limits on the
    retention of the data associated with the instances (e.g., were individuals
    in question told that their data would be retained for a fixed period of
    time and then deleted)?
    }
    If so, please describe these limits and explain how they will be enforced.
    } \\
    No. \\

    \textcolor{datasheet}{\textbf{
    Will older versions of the dataset continue to be
    supported/hosted/maintained?
    }
    If so, please describe how. If not, please describe how its obsolescence
    will be communicated to users.
    } \\
    We will use a versioning system to keep track of the changes in the annotations. Older versions of annotations will be available for download to ensure reproducibility. In case of detected privacy concerns, we will update the image data accordingly. In this case, older version of the data will not be available for download. \\

    \textcolor{datasheet}{\textbf{
    If others want to extend/augment/build on/contribute to the dataset, is
    there a mechanism for them to do so?
    }
    If so, please provide a description. Will these contributions be
    validated/verified? If so, please describe how. If not, why not? Is there a
    process for communicating/distributing these contributions to other users?
    If so, please provide a description.
    } \\
    We encourage the community to expore other uses of the dataset and extend it with new types of annotations. The users creating the new annotations will be responsible for hosting and distributing their annotations. \\

    \textcolor{datasheet}{\textbf{
    Any other comments?
    }} \\
    No. \\

}

\end{document}